\documentclass[journal]{IEEEtran}

\IEEEoverridecommandlockouts

\usepackage[utf8]{inputenc}

\newcommand\approachName{Parallel Extraction for Long-Horizon Disassembly\xspace}
\newcommand\approachNameHighlight{\textbf{P}arall\textbf{E}l \textbf{E}xtraction for \textbf{L}ong-Horizon Disassembly\xspace}
\newcommand\approachNameAbbrv{PEEL\xspace}

\usepackage{xcolor}

\usepackage{balance}
\usepackage{amsmath}
\usepackage{amsfonts}
\usepackage[ruled,vlined,linesnumbered]{algorithm2e}
\usepackage{pgffor}
\usepackage{multirow}
\usepackage{graphicx}
\usepackage{tabularx}
\DontPrintSemicolon

\SetAlFnt{\small}
\usepackage{float}

\usepackage{adjustbox}
\newcommand\cbox[1]{\raisebox{0.1cm}{\colorbox{#1}{}}}

\definecolor{cmab}{HTML}{E74C3C}
\definecolor{crrt}{HTML}{3498DB}
\definecolor{ctrrt}{HTML}{9B59B6}                   
\definecolor{cmatevec}{HTML}{E67E22}

\usepackage{booktabs}
\usepackage{array}
\usepackage{tikz}
\usetikzlibrary{shapes.geometric, arrows.meta, positioning, fit, backgrounds, calc, decorations.pathreplacing}

\usepackage{graphicx}
\usepackage[font=small, labelfont=bf]{caption}
\usepackage{subcaption}

\usepackage[
activate   = {true},
protrusion = false,
expansion  = true,
kerning    = true,
spacing    = true,
tracking   = false,
auto       = true,
selected   = true,
factor     = 1000,
stretch    = 10,
shrink     = 10,
]{microtype}

\makeatletter

\renewcommand\paragraph{\@startsection{paragraph}{4}{\z@}%
                                    {0.2em}%
                                    {0em}%
                                    {\hspace{-0.5em}\noindent\bfseries\normalsize}}
\makeatother

\makeatletter
\let\NAT@parse\undefined
\makeatother
\definecolor{hrefcolor}{HTML}{2c778f}
\usepackage[
pdfa,
colorlinks,
bookmarksopen,
bookmarksnumbered,
allcolors=hrefcolor
]{hyperref}
\usepackage[english]{babel}
\usepackage{amsthm}

\newtheorem{proposition}{Proposition}

\usepackage[nameinlink,capitalise]{cleveref}
\crefname{lemma}{Lemma}{Lemmas}
\Crefname{lemma}{Lemma}{Lemmas}
\crefname{corollary}{Corollary}{Corollaries}
\Crefname{corollary}{Corollary}{Corollaries}
\crefname{proposition}{Proposition}{Propositions}
\Crefname{proposition}{Proposition}{Propositions}
\crefname{line}{line}{lines}
\crefname{figure}{Fig.}{Figs.}
\Crefname{figure}{Fig.}{Figs.}
\crefname{equation}{Eq.}{Eqs.}
\Crefname{equation}{Eq.}{Eqs.}
\crefname{section}{Sec.}{Secs.}
\Crefname{section}{Sec.}{Secs.}
\crefname{definition}{Def.}{Defs.}
\Crefname{definition}{Def.}{Defs.}
\crefname{algorithm}{Alg.}{Algs.}
\Crefname{algorithm}{Alg.}{Algs.}
\crefname{table}{Tbl.}{Tbls.}
\Crefname{table}{Tbl.}{Tbls.}

\author{Servet B. Bayraktar and Andreas Orthey and Zachary Kingston and Marc Toussaint}
\title{\Huge Large-Scale Disassembly: Efficient Planning for
Recycling using Path Defragmentation }
\title{\Huge Multi-Part Disassembly using Scale-Invariant Sampling}
\title{\Huge \approachName: Long-Horizon Multi-Part\\Disassembly Planning via Scale-Invariant Sampling}
\title{\Huge PEEL: Parallel Extraction for Long-Horizon Disassembly Planning via Scale-Invariant Sampling}
\begin{document}
\maketitle
\begin{abstract}
Long-horizon multi-part object disassembly requires robots to compute feasible sequences of collision-free removal motions, even in the presence of tight, narrow escape corridors. 
To efficiently solve such disassembly problems, we propose \approachName (\approachNameAbbrv), an algorithm which efficiently computes disassembly motions for object assemblies and feeds them to a robot manipulator for execution. 
\approachNameAbbrv uses sampling-based motion planning to compute single-object motions through the use of a scale-invariant sampling scheme, where the object scale is estimated in a burn-in phase and a subsequent directional sampler exploits the scale. 
This sampling scheme is integrated into a multi-arm bandit rapidly-exploring random tree (MAB-RRT) planner, which switches between different samplers depending on the reward signal received. 
Using MAB-RRT, the \approachNameAbbrv algorithm runs a batch of planners in parallel to obtain an ordered graph specifying the sequence in which object parts have to be removed.
We show that MAB-RRT can efficiently solve single-part disassemblies with 100 percent success rate on 76 assemblies, and that it is robust to its parameters.
By integrating MAB-RRT into \approachNameAbbrv, we solve four long-horizon disassembly problems using the Fetch manipulator robot involving 10 to 17 individual object parts. Further animations, code, and videos can be found at \url{https://peel-disassembly.surge.sh/}.

\end{abstract}

\section{Introduction}

\begin{figure}
    \centering   
    \includegraphics[width=0.43\linewidth]{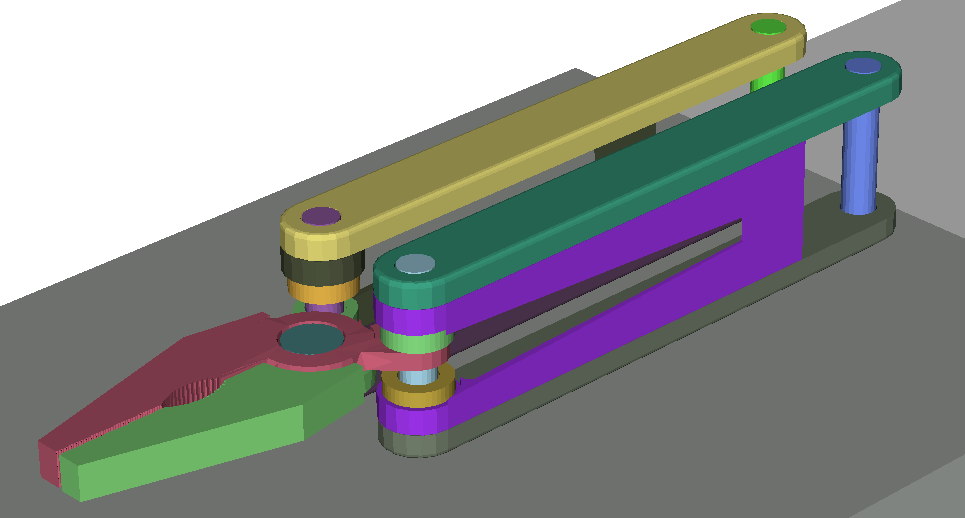}
    \includegraphics[width=0.55\linewidth]
    {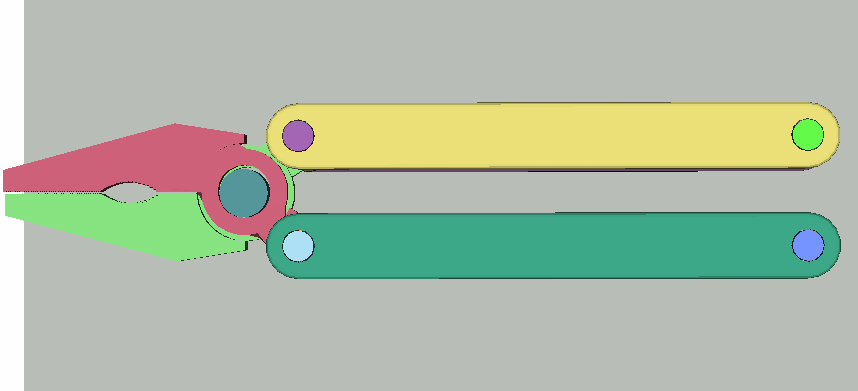}
    %%%
    %\includegraphics[width=0.2\linewidth]{images/pullfigure/Screenshot_at_16-54-40.png}
    \includegraphics[width=\linewidth]
    {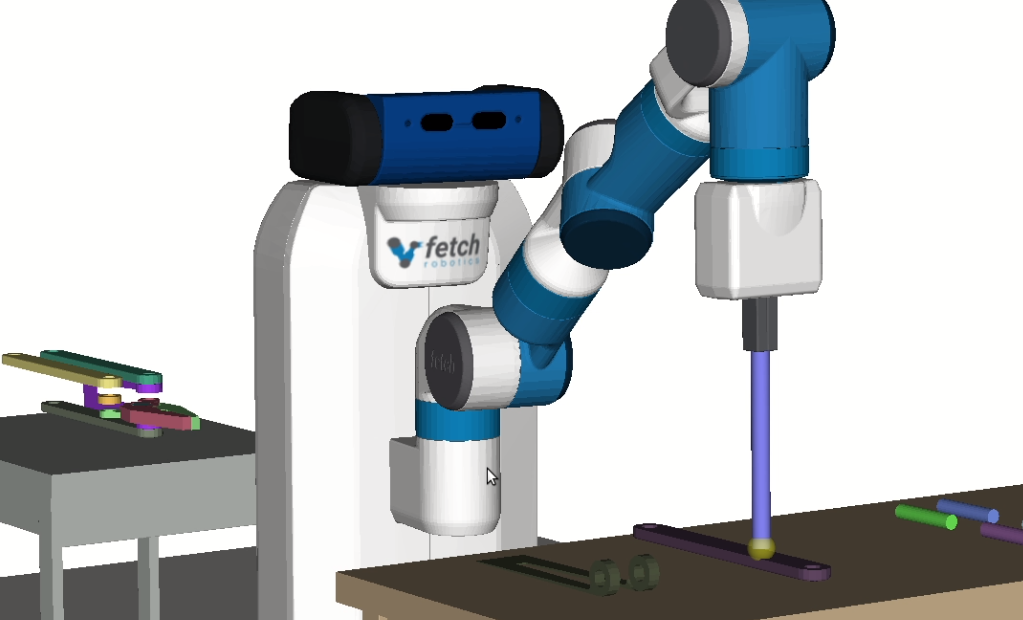}
    %%%
    \includegraphics[width=\linewidth]{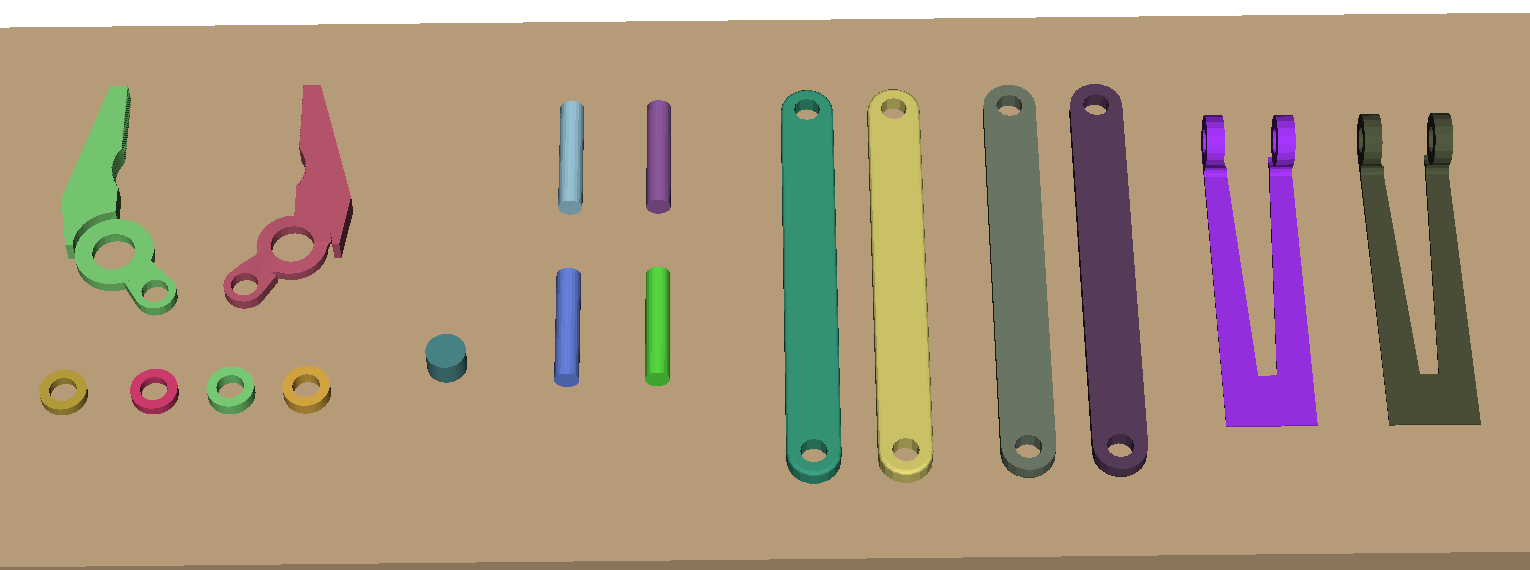}
    \caption{A Fetch mobile manipulator executing a disassembly sequence on a pair of pliers. Top: A pair of pliers in its assembled state viewed from its side (left) and from the top (right). Middle: the robot places a disassembled part at its goal location on a table. Bottom: The disassembled pieces are placed side-by-side on the table.}
    \label{fig:page1}
\end{figure}

Multi-part object disassembly is a fundamental robot skill with applications in recycling, remanufacturing, repair, material recovery, and building deconstruction. 
Disassembling arbitrary objects requires detecting and identifying parts,
removing them without damaging the assembly, and computing action sequences that
systematically dismantle an object.
A core challenge hereby is disassembly planning, where we need to determine the order in which parts should be removed while ensuring geometric feasibility.

Prior approaches to disassembly planning have primarily relied on symbolic or graph-based search over part precedence relations \cite{homemdemello1990andor}. 
While effective for small assemblies, these methods scale poorly as exhaustive
graph search grows exponentially with part count. Physics-based planners~\cite{tian2022assemble} capture
contacts and forces in detail but are computationally expensive for large
assemblies. Geometric planners abstract away physical forces and consider only
collision-free motion. 
While this abstraction may yield solutions requiring refinement for physical
execution, geometric planners produce feasible candidates quickly and provide a
foundation for subsequent physics-aware reasoning~\cite{tian2022assemble}.

In this work, we propose an efficient geometric motion planning method that analyzes a given assembly, computes a collision-free disassembly sequence, and generates motion plans which can be executed on a robot platform.

% Parallel batched execution flow figure for thesis Chapter 4
% (Algorithm 4: parallel batched execution).
%
% Layout grid (cm; all coordinates are box CENTERS unless noted):
%   Queue Q       : (0.0, -2.0), 1.7 x 4.6
%   Fan-out rail  : vertical line at x = 3.3
%   Planner p1    : (6.0,  0.0), 2.0 x 1.1
%   Planner p2    : (6.0, -1.5), 2.0 x 1.1
%   Planner pB    : (6.0, -4.0), 2.0 x 1.1
%   Concurrent bg : (6.0, -2.0), 3.0 x 5.9
%   Race diamond  : (10.0, -2.0), 2.4 x 1.7
%   Winner box    : (13.5, 0.0), 2.4 x 1.0
%   Sequence S    : (17.5, 0.0), 3.2 x 1.4
%   Requeue box   : (17.5, -4.0), 3.4 x 1.2

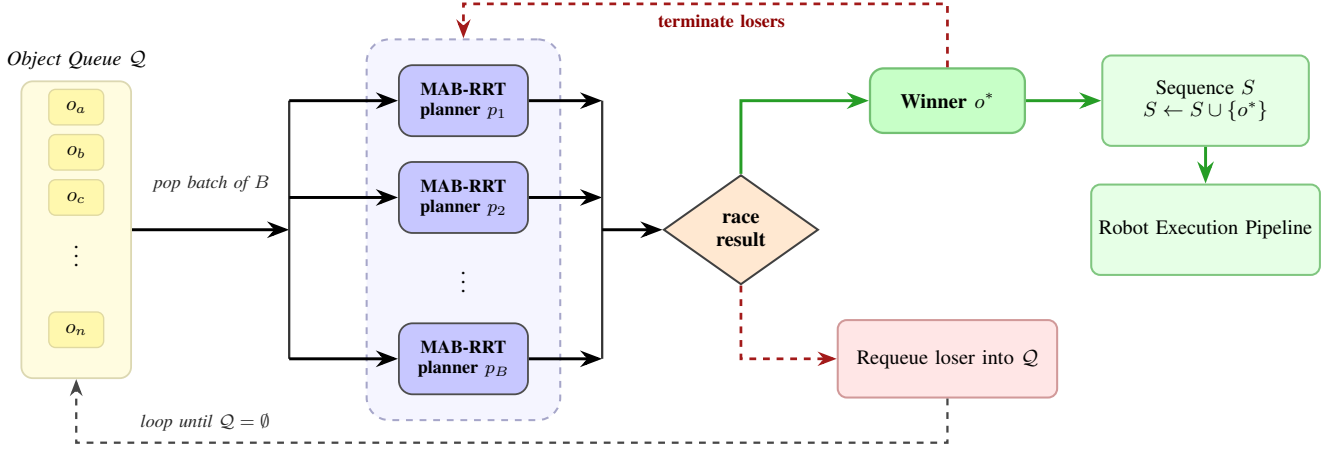
\begin{figure*}
\centering
\resizebox{0.97\linewidth}{!}
{%
\begin{tikzpicture}[
    queue/.style       ={rectangle, rounded corners=3pt, draw=yellow!70!black!50,
                         line width=0.6pt, fill=yellow!40,
                         minimum width=0.85cm, minimum height=0.55cm,
                         align=center, font=\small\bfseries},
    queuebox/.style    ={rectangle, rounded corners=5pt, draw=yellow!60!black!40,
                         line width=0.8pt, fill=yellow!15, inner sep=5pt},
    planner/.style     ={rectangle, rounded corners=6pt, draw=black!70,
                         line width=0.8pt, fill=blue!22,
                         minimum width=2.0cm, minimum height=1.1cm,
                         align=center, font=\footnotesize\bfseries},
    racepoint/.style   ={diamond, draw=black!75, line width=0.9pt, fill=orange!18,
                         aspect=1.4, inner sep=0pt,
                         minimum width=2.4cm, minimum height=1.7cm,
                         align=center, font=\small\bfseries},
    winnerbox/.style   ={rectangle, rounded corners=6pt, draw=green!55!black!55,
                         line width=1.0pt, fill=green!22,
                         minimum width=2.4cm, minimum height=1.0cm,
                         align=center, font=\small\bfseries},
    seqbox/.style      ={rectangle, rounded corners=4pt, draw=green!55!black!50,
                         line width=0.9pt, fill=green!10,
                         minimum width=3.2cm, minimum height=1.4cm,
                         align=center, font=\small},
    requeuebox/.style  ={rectangle, rounded corners=4pt, draw=red!55!black!40,
                         line width=0.9pt, fill=red!10,
                         minimum width=3.4cm, minimum height=1.2cm,
                         align=center, font=\small},
    concurrentbg/.style={rectangle, rounded corners=10pt, draw=blue!40!black!35,
                         dashed, line width=0.9pt, fill=blue!4},
    arrow/.style       ={-{Stealth[scale=1.1]}, line width=1.3pt},
    succarrow/.style   ={-{Stealth[scale=1.1]}, green!55!black!85, line width=1.3pt},
    killarrow/.style   ={-{Stealth[scale=1.0]}, red!60!black!90,
                         line width=1.2pt, dashed},
    looparrow/.style   ={-{Stealth[scale=1.0]}, gray!55!black,
                         line width=1.0pt, dashed},
    feedline/.style    ={-, gray!40!black, line width=0.9pt},
    annot/.style       ={font=\footnotesize\itshape, gray!30!black, inner sep=1pt},
    succannot/.style   ={font=\footnotesize\itshape, green!45!black, inner sep=1pt},
    killannot/.style   ={font=\footnotesize\bfseries, red!55!black, inner sep=1pt},
    titleannot/.style  ={font=\small\itshape, blue!40!black, inner sep=1pt},
]

% =================================================================
% (1) QUEUE Q
% =================================================================
\node[queuebox, minimum width=1.7cm, minimum height=4.6cm]
     (qbox) at (0.0, -2.0) {};
\node[above=0.05cm of qbox, font=\small\itshape] {Object Queue $\mathcal{Q}$};

\node[queue] at (0.0, -0.10) {$o_a$};
\node[queue] at (0.0, -0.80) {$o_b$};
\node[queue] at (0.0, -1.50) {$o_c$};
\node[font=\normalsize] at (0.0, -2.30) {$\vdots$};
\node[queue] at (0.0, -3.55) {$o_n$};

% =================================================================
% (2) CONCURRENT REGION + PLANNER STACK
%     Planners 2.0 wide -> west=5.0, east=7.0
%     Concbg 3.0 wide   -> west=4.5, east=7.5
% =================================================================
\node[concurrentbg, minimum width=3.0cm, minimum height=5.9cm]
     (concbg) at (6.0, -2.0) {};
%\node[titleannot, anchor=south west, align=left]
%   at ($(concbg.north west) + (-1.6, 0.10)$)
%   {Parallel Execution};

\node[planner] (p1) at (6.0,  0.00) {MAB-RRT\\planner $p_1$};
\node[planner] (p2) at (6.0, -1.50) {MAB-RRT\\planner $p_2$};
\node[font=\normalsize] (pdots) at (6.0, -2.70) {$\vdots$};
\node[planner] (pB) at (6.0, -4.00) {MAB-RRT\\planner $p_B$};

% =================================================================
% (3) POP-AND-FAN-OUT  (rail at x=3.3, outside concbg.west=4.5)
% =================================================================
\coordinate (railTop) at (3.3,  0.00);
\coordinate (railMid) at (3.3, -2.00);
\coordinate (railBot) at (3.3, -4.00);

\draw[arrow] (qbox.east) -- (railMid);
\node[annot, anchor=south] at (2.1, -1.45) {pop batch of $B$};

\draw[feedline, line width=1.0pt] (railTop) -- (railBot);
\draw[arrow] (railTop)    -- (p1.west);
\draw[arrow] (3.3, -1.50) -- (p2.west);
\draw[arrow] (railBot)    -- (pB.west);

% =================================================================
% (4) RACE diamond
% =================================================================
\node[racepoint] (race) at (10.3, -2.0) {race\\result};

\coordinate (endTop) at (8.15,  0.00);
\coordinate (endMid) at (8.15,  -2.00);
\coordinate (endBot) at (8.15,  -4.00);

\draw[arrow] (p1.east) -- (endTop);
\draw[arrow] (p2.east) -- (8.15,-1.50);
\draw[arrow] (pB.east) -- (endBot);
\draw[feedline, line width=1.0pt] (endTop) -- (endBot);
\draw[arrow] (endMid) -- (race.west);

% p1 solid (winner); p2 and pB faded dashed (losers)
%\draw[feedline, -{Stealth[scale=0.85]}] (p1.east) -- (race.west);
%\draw[gray!25, line width=0.7pt, dashed, -{Stealth[scale=0.75]}] (p2.east) -- (race.west);
%\draw[gray!25, line width=0.7pt, dashed, -{Stealth[scale=0.75]}] (pB.east) -- (race.west);

% =================================================================
% (5) WINNER + SEQUENCE (success path)
% =================================================================
\node[winnerbox] (winner) at (13.5,  0.0) {Winner $o^{*}$};
\node[seqbox]    (seq)    at (17.5,  0.0)
     {Sequence $S$\\$S \leftarrow S \cup \{o^{*}\}$};

\draw[succarrow] (race.north) |- (winner.west);

\draw[succarrow] (winner.east) -- (seq.west);

\node[seqbox]    (exec)    at (17.5,  -2.0)
     {Robot Execution Pipeline};
\draw[succarrow] (seq.south) -- (exec.north);

% =================================================================
% (6) WINNER triggers SIGKILL on the BATCH (concurrent block)
% =================================================================
\node[requeuebox] (requeue) at (13.5, -4.0)
     {Requeue loser into $\mathcal{Q}$};

\draw[killarrow] (winner.north) -- (winner.north |- 0,+1.5)
    -- (concbg.north |- +0.0,+1.5) -- (concbg.north);
\node[killannot, anchor=south] at (10,1.1)
     {terminate losers};

% =================================================================
% (7) Timeout wall: standalone vertical line in the gap
%     between concbg.east (x=7.5) and race.west (x=8.8)
% =================================================================
%\draw[line width=1.2pt, dashed, red!60!black]
%    (8.15, 0.8) -- (8.15, -4.8);
%\node[killannot, anchor=north, font=\scriptsize\itshape]
%    at (8.65, -4.85)
%    {timeout $T_{\text{batch}}$};

% =================================================================
% (8) Race-result -> Requeue
% =================================================================
\draw[killarrow] (race.south) |- (requeue.west);

% =================================================================
% (9) LOOP-BACK rail
% =================================================================
\draw[looparrow]
   (requeue.south)
   -- (requeue.south |- 0,-5.3)
   -- (0.0, -5.3)
   -- (qbox.south);
\node[annot, anchor=north] at (2, -4.8) {loop until $\mathcal{Q} = \emptyset$};

\end{tikzpicture}
}
\caption{Parallel batch protocol (Algorithm~\ref{alg:batched}).
At each round, up to $B$ objects are popped from the shuffled queue
$\mathcal{Q}$ and assigned to concurrent MAB-RRT planner processes that
share the current planning budget $T$.
The race resolves into one of two cases.
\emph{(Success)} The first planner to finish yields the winner $o^{*}$,
which is committed to the solution sequence $S$;
all remaining batch processes (the losers) are terminated,
their objects are returned to $\mathcal{Q}$, and $T$ is reset to $T_0$ because the geometry has changed.
\emph{(Timeout)} If no planner finishes before $T$ elapses,
all processes are terminated and the full batch is requeued. Once every unsolved object has been attempted at the current budget, that is, once one full sweep has failed, $T$ is doubled.
The procedure repeats until $\mathcal{Q}$ is empty. The output is a sequence graph $S$, which determines the order of objects and object paths, which can then be used in robot execution.}
\label{fig:parallel_batched_execution}
\end{figure*}

The core difficulty in geometric disassembly differs fundamentally from typical motion planning. Rather than exploring a broad configuration space, the planner must identify narrow escape corridors to move each part from its constrained initial state into free space. For example, to disassemble a pair of pliers (Fig.~\ref{fig:page1}), bolts have to be removed from a thin corridor within an otherwise blocked configuration space. Uniform sampling rarely discovers such corridors~\cite{salzman2014power}, and naive directional sampling~\cite{dalibard2011linear} can become inefficient without proper initialization.
%
%To efficiently solve disassembly sequences amid narrow escape corridors while avoiding exponential growth in the number of subassemblies, we introduce the \approachName (\approachNameAbbrv). \approachNameAbbrv is a high-level algorithm which relies at its core on a disassembly-based motion planner. In this work, we utilize the multi-arm bandit RRT (MAB-RRT)~\cite{bayraktar2026wafr} planner, which uses scale-invariant sampling to quickly extract a single part from the surrounding assembly. To solve multi-part assemblies with \approachNameAbbrv, we execute parallel races where multiple candidate objects are planned concurrently. The first to succeed is removed from the assembly, all other planners are terminated, and unsolved objects are requeued for future batches. This winner-based strategy avoids combinatorial explosion and adapts naturally as objects are removed.

Beyond extracting a single part, multi-part disassembly requires deciding the order in which parts are removed. The number of reachable subassembly states grows exponentially with the part count, so explicitly searching over part precedence relations~\cite{homemdemello1990andor} quickly becomes intractable. 
We observe, however, that whether a part is currently removable is decided by the very same query a motion planner already answers: does a collision-free extraction motion exist? This lets us sidestep the symbolic ordering search and allows geometric feasibility itself to reveal the removal order.

Building on this observation, we introduce the \approachNameHighlight (\approachNameAbbrv). Rather than committing to a fixed order, \approachNameAbbrv plans the removal of all currently blocked candidate parts concurrently in a batch, using the multi-arm bandit RRT (MAB-RRT)~\cite{bayraktar2026wafr} planner as the underlying single-object planner, which uses scale-invariant sampling to quickly extract a part from its surrounding assembly. 
The first planner to find a valid escape motion wins: its part is removed, all competing planners are terminated, and any unsolved parts are requeued for the next batch on the reduced assembly. 
This winner-based race avoids combinatorial subassembly enumeration, exploits parallel compute, and adapts naturally as the assembly is dismantled.

To ensure that MAB-RRT works reliably as a single-object planner in \approachNameAbbrv, we benchmark MAB-RRT on 76 assemblies from the Automate dataset \cite{tang2024automate}. On this benchmark, MAB-RRT achieves a 100\% success rate under strict collision checking, outperforming baseline geometric planners~\cite{tian2022assemble}. Additionally, we evaluate its hyperparameters and demonstrate robustness across parameter settings. By using MAB-RRT inside \approachNameAbbrv, we validate \approachNameAbbrv on four multi-part assemblies containing 10–17 parts and demonstrate execution on a robot manipulator for all four assemblies.

Our contributions are, therefore, as follows:
\begin{itemize}
    \item Introduction of the \approachName (\approachNameAbbrv), a greedy framework for multi-part disassembly that replaces the search over part precedence relations with concurrent planning races: candidate parts are planned concurrently in batches with MAB-RRT~\cite{bayraktar2026wafr}.
    \item Extensive benchmarking of MAB-RRT on 76 benchmark assemblies shows robustness to hyperparameters and improved runtime performance over baseline methods.
    \item Integration of the results of \approachNameAbbrv into a robot execution pipeline that computes collision-free manipulation actions for a robot manipulator, including possible regrasping actions in case of failure.  
    \item Demonstration of robot execution in a physical simulation on four multi-part assemblies by using the computed precedence graphs. 
\end{itemize}
\section{Related Work}

This work is closely related to disassembly sequence planning,
object extraction, and parallel motion planning. We review those topics below and discuss their relation to our work.
%%%%%%%%%%%%%%%%%%%%%%%%%%%%%%%%%%%%%%%%%%%%%%%%%%%%%%%%%%%%%%%%%%%%%%%%%%%%%%%%%%%%%%%
\subsection{Disassembly Sequence Planning\label{sec:rw_disassembly_sequence}}
%%%%%%%%%%%%%%%%%%%%%%%%%%%%%%%%%%%%%%%%%%%%%%%%%%%%%%%%%%%%%%%%%%%%%%%%%%%%%%%%%%%%%%%

Disassembly planning~\cite{asif2024robotic} has historically been treated
as a combinatorial sequencing problem. The standard representation is the
AND/OR graph~\cite{homemdemello1990andor,homemdemello1991correct}, whose
nodes are subassemblies and whose hyperarcs are binary partitions that can
be separated. Such graphs represent the assembly space completely, but the
number of candidate partitions grows exponentially with the part
count~\cite{wilson1994geometric}, and building one from CAD data exhibits
$n!$-proportionate computational behavior~\cite{munker2022cad}. These costs
are accumulated before any sequence can be selected, which makes the approach
impractical beyond roughly 20 parts~\cite{munker2022cad}.

Sequencing methods differ in the type of feasibility they establish.
Symbolic feasibility encodes part relationships in precedence
graphs~\cite{niu2003hierarchical,wang2016weighted} or interference matrices
and treats each removal as atomic. Wilson and
Latombe~\cite{wilson1994geometric} showed that deciding whether a
subassembly can separate requires geometric reasoning about blocking relationships in
every direction of motion. Their non-directional blocking graph captures
these constraints in polynomial time, but tests only infinitesimal motions
and therefore does not certify global removability. Geometric feasibility
instead integrates path planning into sequencing to verify that a removal
is realizable~\cite{sundaram2001disassembly}, though such methods still
rely on a precomputed precedence structure to generate the
candidates~\cite{jcortesDisassembly2010}.

Sequencing methods also differ in when they declare a part separated. The strongest notion requires the part to reach the unbounded free-space component, that is, to be removable to infinity \cite{wilson1994geometric}. Most systems instead check a cheaper surrogate, such as convex-hull separation \cite{tian2022assemble}, exit from a bounding box \cite{aguinaga2008targetless}, or a displacement threshold \cite{tian2024asap}, none of which certifies global separability once rotation is allowed.

The most directly comparable multi-part system is AssembleThemAll
(ATA)~\cite{tian2022assemble}, which casts assembly as disassembly inside a
physics simulator: the continuous six-dimensional motion search is replaced
by a breadth-first search over a discrete action space of forces applied
for fixed time steps, and contacts are resolved by \emph{signed distance
field} (SDF) collision detection under a non-zero penetration threshold,
since convex-hull decomposition suits neither complex concave geometry nor
the small inherent overlaps common in CAD assemblies. Its sequencing layer
is a progressive breadth-first search over parts: each iteration attempts
every remaining part at a bounded search depth, commits the first that
succeeds, and increments the bound until all parts are removed.
ASAP~\cite{tian2024asap} extends ATA with gravitational stability, robot
integration, and a learned part-selection policy, requiring each candidate
subassembly state to be both reachable by a path planner and statically
stable under gravity.

\paragraph{Learning-based ordering}

A complementary line of work attacks the combinatorial cost of sequencing
by learning which parts are likely removable, ordering the removal-test
queue by a learned prior instead of searching it exhaustively. Cebulla et
al.~\cite{cebulla2023speeding} predict per-part removability with a graph
neural network over contact-graph representations of CAD assemblies, reducing
the number of removal tests by 64\% to 90\% on five real-world assemblies of
bolted aluminium framing, of 29 to 68 parts and so larger than the assemblies we
evaluate. The learned policy of ASAP~\cite{tian2024asap} serves the same
purpose.

\paragraph{Parallel disassembly work}
Parallelism has been applied to disassembly before, though not to sequence
discovery. Aguinaga et al.~\cite{aguinaga2008targetless} parallelize
RRT-based path planning for selective disassembly, while Ren et
al.~\cite{ren2018asynchronous} and Zhang et al.~\cite{zhang2014parallel}
parallelize execution, respectively scheduling concurrent manipulators
under precedence and collision constraints and identifying groups of parts
removable simultaneously. Dedicated part-removal query
planners~\cite{zhang2008d} and broader metaheuristic
surveys~\cite{guo2021dsp} complete the picture.

Across these approaches a common pattern emerges: sequencing and the
geometric feasibility check are treated as separate stages. A precedence
structure is computed first, whether by graph search, interference
matrices, learned priors, or progressive search over discrete actions, and
motion planning, if performed at all, only validates the candidates that
layer proposes. \approachNameAbbrv removes the separation, since each round
of its parallel race is a set of per-part motion planning queries whose
winner defines the next removal, so no precedence structure is ever
constructed. We also check collisions with bounding volume hierarchies at
zero penetration tolerance rather than SDF detection under a non-zero
threshold~\cite{tian2022assemble}, which restricts the admissible input,
as we filter assemblies for strict separation, but removes any reliance on
a penetration tolerance.

%%%%%%%%%%%%%%%%%%%%%%%%%%%%%%%%%%%%%%%%%%%%%%%%%%%%%%%%%%%%%%%%%%%%%%%%%%%%%%%%%%%%%%%
\subsection{Object Extraction}\label{sec:rw_object_extraction}
%%%%%%%%%%%%%%%%%%%%%%%%%%%%%%%%%%%%%%%%%%%%%%%%%%%%%%%%%%%%%%%%%%%%%%%%%%%%%%%%%%%%%%%
Object extraction, the separation of one object from its surroundings, is
the operation on which any disassembly sequence rests. Typically the object
is a bolt, screw, nut, or gear that has to be moved through an elongated
narrow passage~\cite{tian2022assemble}. Special cases admit shortcuts:
screw removal reduces largely to choosing the correct
tool~\cite{zhang2023automatic,li2020unfastening}, cluttered environments
offer wider passages where generic sampling-based methods
suffice~\cite{Bayraktar2023RAL}, and objects piled on each other require
methods that keep the pile from
collapsing~\cite{pathak2025collapse,Motoda2023}. We target the general
case, where narrow passages occur for almost every part.

Two frameworks have proven particularly effective. The first is
physics-based simulation~\cite{tian2022assemble,zickler2009efficient},
which applies random forces to the object; although most point in the wrong
direction, their projection onto the correct escape direction still
advances it~\cite{tian2022assemble}. Behavioral Kinodynamic RRT
(BK-RRT)~\cite{zickler2009efficient} treats such forces as random controls,
as kinodynamic systems are moved, while ATA's disassembly breadth-first
search~\cite{tian2022assemble} explores reachable states with a similarity
check that avoids duplicate directions. However, such simulation is 
costly~\cite{tian2022assemble}.

The second framework is biased sampling in sampling-based motion
planning~\cite{Orthey2023AnnualReview}, which avoids simulating dynamics.
Samplers bias toward object
boundaries~\cite{Boor1999GaussianSampling,Amato1998ObstacleBased} or
directly toward narrow passages~\cite{Hsu2003BridgeTest}; utility-based
sampling scores samples by their expected contribution to planning
progress~\cite{burns2005toward}; and the dynamic-domain
RRT~\cite{yershova2005dynamic} adapts a per-node radius to each sample's
success in extending the tree, which helps overcome narrow passages and even
explore zero-measure manifolds for contact
planning~\cite{yershova2009motion}.

Biased sampling can also be specialized to extraction. Manhattan-like
RRT~\cite{cortes2008disassembly} plans for the object under consideration
and moves the remaining degrees of freedom only when they block the
removal. The same decomposition appears in factored state
spaces~\cite{Bayraktar2023RAL}, where interpolating one factor at a time within
a single tree advances one object while the others stay passive. Targetless-RRT~\cite{aguinaga2008targetless} replaces an explicit
target with an implicit goal region, such as a separation threshold between
object and environment. Mating Vector RRT
(MateVec-TRRT)~\cite{Ebinger2018MateVecTRRT} computes mating vectors along
which separation from the environment is likely to increase. Most recently,
the classical multi-arm bandit
(MAB)~\cite{auer2002finite,bubeck2012regret,slivkins2019introduction} has
been combined with RRT into
MAB-RRT~\cite{faroni2023motion,faroni2024online,bayraktar2026wafr}, letting
a bandit arbitrate online between samplers and so admitting dedicated
disassembly samplers~\cite{bayraktar2026wafr}: scale-invariant sampling
first locates the scale at which the free space around the object is best
resolved, and a principal component analysis
(PCA)~\cite{dalibard2009control,dalibard2011linear} of the samples taken at
that scale then gives the direction along which to advance locally.

We use MAB-RRT~\cite{bayraktar2026wafr} directly as the single-object
planner inside \approachNameAbbrv (\cref{sec:scale_sampler}); our
contribution is the protocol built around it, which targets full multi-part
disassembly rather than single-object extraction. To establish it as a
reliable component in this setting, we benchmark it under strict collision
checking and study its parameter sensitivity
(\cref{sec:single_part_benchmarks,sec:sensitivity}).

%%%%%%%%%%%%%%%%%%%%%%%%%%%%%%%%%%%%%%%%%%%%%%%%%%%%%%%%%%%%%%%%%%%%%%%%%%%%%%%%%%%%%%%
\subsection{Parallel and Distributed Motion Planning}\label{sec:rw_parallel_planning}
%%%%%%%%%%%%%%%%%%%%%%%%%%%%%%%%%%%%%%%%%%%%%%%%%%%%%%%%%%%%%%%%%%%%%%%%%%%%%%%%%%%%%%%

Parallel computation has been applied to motion planning at two levels.
Within a single query, planning primitives such as collision checking and
nearest-neighbor search are parallelized: PRRT and
PRRT*~\cite{ichnowski2014scalable} build one shared RRT on multicore CPUs
using lock-free concurrency and partition-based sampling,
pRRTC~\cite{huang2025prrtc} runs hundreds of concurrent RRT-Connect
iterations on a GPU over shared start and goal trees, and
cuRobo~\cite{sundaralingam2023curobo} batches trajectory optimization
across seeds and keeps the best. Across queries, independent planners
cooperate: C-FOREST~\cite{otte2013cforest} shares improving solutions
between trees to enable pruning and tighter sampling bounds, an advantage
that diminishes for feasibility problems where the first solution suffices;
distributed RRTs partition the configuration space across processors and
merge the resulting subtrees~\cite{jacobs2013scalable}; SRT grows
independent local trees in parallel and connects them into a
roadmap~\cite{plaku2005srt}; and dRRT searches the implicit tensor product
of independently built per-robot roadmaps~\cite{shome2020drrt}.

All of these parallelize a single planning query. \approachNameAbbrv
instead parallelizes across \emph{candidate parts}, so the race between
concurrent planners is what determines the removal order. This
winner-takes-all scheme is a form of
OR-parallelism~\cite{ichnowski2014scalable, otte2013cforest}, applied to
discover disassembly sequences rather than to accelerate a single plan. The
decomposition also keeps every query low-dimensional: each planner searches
$\mathrm{SE}(3)$ for one object on a single core, rather than the composite
space of all objects jointly, where sampling-based planners become intractable
and partitioning into independent subproblems would otherwise require
structured decompositions such as factored state
spaces~\cite{Bayraktar2023RAL}. \approachNameAbbrv needs neither.

\section{Multi-Part Disassembly Problem}\label{sec:problem_formulation}

We formulate multi-part disassembly as a sequential geometric motion planning problem. Let an assembly be a set of $N$ rigid objects $\{o_1, o_2, \dots, o_N\}$, each occupying the six-degree-of-freedom configuration space $\mathrm{SE}(3)$. Each object $o_i$ begins at an initial configuration $\mathbf{q}_{i}^{\text{init}}$ and must reach a goal region $G_i$ while avoiding collisions with all other objects and static obstacles. Objects can only be moved by a robot $R$, with configuration space $C$, that has an end-effector to establish a rigid connection between robot and object which is modeled using a welding joint. 

Our formulation is purely geometric: objects are treated as rigid bodies, and feasibility is determined solely by collision constraints. Physical effects such as friction, gravity, and contact forces are not modeled. This abstraction isolates the geometric structure of disassembly planning and focuses on the geometric feasibility problem.  

The sequential nature of disassembly induces strong dependencies between planning problems. Removing one object reduces the collision constraints on the remainder, often enabling motions that were previously blocked. The problem is therefore to identify a sequence of collision-free motions that incrementally dismantles the assembly and executes the sequence using robot $R$.

Let us define this more formally. Let $X \subseteq \{o_1, \dots, o_N\}$ denote all objects that have not yet been removed. An
object $o_i \in X$ is \emph{removable in $X$} if a collision-free path exists
for $o_i$ from $\mathbf{q}_{i}^{\text{init}}$ to $G_i$ treating
$X \setminus \{o_i\}$ as static obstacles, and $X$ is \emph{sequentially
decomposable} if its objects admit an ordering
$o_{\pi(1)}, \dots, o_{\pi(m)}$ in which every $o_{\pi(k)}$ is removable in
$\{o_{\pi(k)}, \dots, o_{\pi(m)}\}$. 
%Such an ordering is \emph{clear} if each of
%its paths has positive clearance. 
We model a removed object as leaving the
scene rather than being displaced within it, which is consistent with placing
every $G_i$ away from the assembly.

This formulation leaves open how the goal region $G_i$ is defined, that is, when a part counts as \emph{disassembled}. As noted in \cref{sec:rw_disassembly_sequence}, the principled criterion, removal to infinity, is expensive to evaluate. We therefore use a computable surrogate: $G_i$ is a region of free space placed away from the assembly, and a part is removed once its position enters $G_i$. Alternatively, an explicit goal configuration can be used. At execution time we detect a local \emph{free state} with a mobility-rank test (\cref{sec:robot_pipeline}). 
%Both are conservative surrogates that depend on tuning rather than guaranteeing global separability. A cheap criterion that certifies true separability, especially under rotation, remains an open problem.    

\section{\approachName\label{sec:parallel_batched}}

To solve the disassembly problem, we propose the \approachName (\approachNameAbbrv) algorithm. \approachNameAbbrv uses, at its core, a single-object planner that employs scale-invariant sampling to separate two parts to extract a single part from the surrounding assembly. Given such a planner, we formulate sequence
discovery as a series of \emph{planning races}: at each round, several
single-object planners execute concurrently, and the first to succeed
determines the winner, after which the round ends.
This race-based strategy incrementally identifies removable objects without building an
explicit search over all orderings. 
\cref{fig:parallel_batched_execution}
illustrates the protocol.

\subsection{High Level Algorithm}

The actual implementation of \approachNameAbbrv is shown as pseudocode in~\cref{alg:batched}. The algorithm starts in Line~\ref{alg:batched:shuffle} by initializing the queue $\mathcal{Q}$ with a random shuffle of all objects; line~\ref{alg:batched:init} initializes
the solution sequence $S$ to empty. Lines~\ref{alg:batched:budgetinit}
and~\ref{alg:batched:sweepinit} initialize the planning budget $T$ to $T_0$ and
the set $\mathcal{F}$ of objects already attempted at that budget to empty. The main loop
(line~\ref{alg:batched:while}) repeats until $\mathcal{Q}$ is empty.
At the start of each iteration, line~\ref{alg:batched:pop} pops up to
$B$ objects from the front of $\mathcal{Q}$ to form the current batch.
Lines~\ref{alg:batched:foreach}--\ref{alg:batched:register} launch one
independent planner process per batch object: each planner treats its
assigned object as movable and all remaining objects
(line~\ref{alg:batched:remaining}) as fixed obstacles.
Line~\ref{alg:batched:wait} blocks until either one process succeeds or
the budget $T$ elapses.

If a winner is found (lines~\ref{alg:batched:ifsuccess}--\ref{alg:batched:requeue_losers}),
its object $o^*$ is committed to $S$
(line~\ref{alg:batched:commit}), all other processes are terminated
(line~\ref{alg:batched:kill}), and their objects are requeued
(line~\ref{alg:batched:requeue_losers}). This winner-takes-all strategy
halts planners working on objects that may become easier to remove once
the winner is extracted. Because the geometry has changed, the budget is
reset to $T_0$ and $\mathcal{F}$ is cleared
(line~\ref{alg:batched:reset}): earlier failures were recorded against an
assembly that no longer exists. 

If instead the budget elapses with no winner
(lines~\ref{alg:batched:timeout_kill}--\ref{alg:batched:giveup}), all
processes are terminated and the entire batch is requeued
(line~\ref{alg:batched:requeue_all}), so that difficult objects do not block
progress indefinitely and instead receive repeated attempts as the geometry
evolves. The batch is added to $\mathcal{F}$
(line~\ref{alg:batched:marksweep}) to record that its objects have now been
attempted at the current budget. Once $\mathcal{F}$ covers the queue
(line~\ref{alg:batched:sweepdone}), one full \emph{sweep} over all unsolved
objects has failed at budget $T$, and only then is the budget doubled and
$\mathcal{F}$ cleared (line~\ref{alg:batched:double}). Escalating per sweep
rather than per batch matters in practice: an object entering a batch for the
first time is never charged an inflated budget that it has not yet been shown
to need. The cap $T_{\max}$ (line~\ref{alg:batched:giveup}) bounds this
escalation so that the algorithm also terminates on assemblies that admit no
sequential disassembly at all.

\begin{algorithm}[t]
  \caption{\approachName}
  \label{alg:batched}
  \DontPrintSemicolon
  \SetAlgoLined
  \LinesNumbered

  \KwIn{%
    Assembly $\mathcal{A}$ with objects $\{o_1, \ldots, o_N\}$,
    batch size $B > 0$,
    initial budget $T_0$,
    budget cap $T_{\max}$%
  }
  \KwOut{Disassembly sequence $S$}

  $\mathcal{Q} \gets \textsc{Shuffle}(\{o_1, \ldots, o_N\})$  \tcp*{Initialize queue}  \label{alg:batched:shuffle}
  $S          \gets \emptyset$                                  \tcp*{Solved sequence}   \label{alg:batched:init}
  $T          \gets T_0$                                        \tcp*{Current planning budget} \label{alg:batched:budgetinit}
  $\mathcal{F}\gets \emptyset$                                  \tcp*{Already attempted at budget $T$} \label{alg:batched:sweepinit}

  \While{$\mathcal{Q} \neq \emptyset$}{ \label{alg:batched:while}

    $\text{batch}     \gets \textsc{PopFront}(\mathcal{Q}, B)$ \tcp*{Up to $B$ objects} \label{alg:batched:pop}
    $\text{processes} \gets \emptyset$\;                                                  \label{alg:batched:preinit}

    \ForEach{$o_i \in \text{batch}$}{ \label{alg:batched:foreach}
      $\text{remaining} \gets \mathcal{Q} \cup (\text{batch} \setminus \{o_i\})$\;                   \label{alg:batched:remaining}
      $p_i              \gets \textsc{LaunchPlanner}(o_i, S, \text{remaining}, T)$\;                  \label{alg:batched:launch}
      $\text{processes} \gets \text{processes} \cup \{(p_i, o_i)\}$\;                                 \label{alg:batched:register}
    }

    $\text{winner} \gets \textsc{WaitForFirstSuccess}(\text{processes}, T)$\; \label{alg:batched:wait}

    \eIf{$\text{winner} \neq \text{null}$}{ \label{alg:batched:ifsuccess}

      $(p^*, o^*) \gets \text{winner}$\;                                                                                     \label{alg:batched:unpack}
      $S          \gets S \cup \{o^*\}$                        \tcp*{Record solved object}             \label{alg:batched:commit}
      \textsc{Terminate}($\{p_i \mid (p_i, o_i) \in \text{processes},\, o_i \neq o^*\}$)             \tcp*{Terminate losers} \label{alg:batched:kill}
      \textsc{Requeue}($\mathcal{Q}, \{o_i \mid (p_i, o_i) \in \text{processes},\, o_i \neq o^*\}$)\;                      \label{alg:batched:requeue_losers}
      $T \gets T_0$, $\mathcal{F} \gets \emptyset$             \tcp*{Geometry changed: reset}         \label{alg:batched:reset}

    }{

      \textsc{Terminate}($\text{processes}$)  \tcp*{Budget elapsed}    \label{alg:batched:timeout_kill}
      \textsc{Requeue}($\mathcal{Q}, \text{batch}$)\;                   \label{alg:batched:requeue_all}
      $\mathcal{F} \gets \mathcal{F} \cup \text{batch}$        \tcp*{Attempted at $T$} \label{alg:batched:marksweep}

      \If{$\mathcal{Q} \subseteq \mathcal{F}$}{ \label{alg:batched:sweepdone}
        $T \gets 2T$, $\mathcal{F} \gets \emptyset$   \tcp*{Sweep failed: escalate} \label{alg:batched:double}
        \lIf{$T > T_{\max}$}{\Return{$S$}}                                           \label{alg:batched:giveup}
      }

    }
  }

  \Return{$S$}\; \label{alg:batched:return}
\end{algorithm}
\begin{figure*}[t]
\centering
\resizebox{\linewidth}{!}
{%
\begin{tikzpicture}[
    % Node styles
    phase/.style={rectangle, rounded corners=6pt, draw=black!70, line width=0.8pt, fill=blue!25,
                  minimum width=2.0cm, minimum height=1.3cm, align=center, font=\small\bfseries},
    currobj/.style={rectangle, rounded corners=4pt, draw=gray!70, line width=0.8pt, fill=gray!15,
                  minimum width=1.6cm, minimum height=1.3cm, align=center, font=\small},
    assembly/.style={rectangle, rounded corners=4pt, draw=gray!70, line width=0.8pt, fill=gray!20,
                  minimum width=2.0cm, minimum height=0.9cm, align=center, font=\small},
    output/.style={rectangle, rounded corners=4pt, draw=yellow!50!black!50, line width=0.6pt, fill=yellow!15,
                  minimum width=1.8cm, minimum height=0.9cm, align=center, font=\small},
    offline/.style={rectangle, rounded corners=6pt, draw=black!70, line width=0.8pt, fill=orange!30,
                   minimum width=2.0cm, minimum height=1.3cm, align=center, font=\small\bfseries},
    queue/.style={rectangle, rounded corners=3pt, draw=green!70!black!50, line width=0.6pt, fill=green!40,
                  minimum width=0.9cm, minimum height=0.55cm, align=center, font=\small\bfseries},
    queuebox/.style={rectangle, rounded corners=5pt, draw=green!60!black!40, line width=0.8pt, fill=green!15,
                     inner sep=5pt},
    dashedbox/.style={rectangle, rounded corners=10pt, draw=black!50, dashed, line width=1.2pt},
    arrow/.style={-{Stealth[scale=1.1]}, line width=1.5pt},
    dataarrow/.style={-{Stealth[scale=0.8]}, gray!70, line width=0.8pt},
    looparrow/.style={-{Stealth[scale=1.0]}, line width=1.2pt, dashed},
]

% Define the main vertical center line
\def\mainY{0}

% ===== OFFLINE SECTION =====
% MAB-RRT Planning - at main height
\node[offline] (mabrrt) at (0, \mainY) {MAB-RRT\\Planning};

% Assembly Model above MAB-RRT
\node[assembly, below=0.5cm of mabrrt] (assembly) {Assembly\\Model};

% Arrow from Assembly to MAB-RRT
\draw[arrow] (assembly) -- (mabrrt);

% ===== QUEUE SECTION =====
\node[queuebox, minimum width=1.3cm, minimum height=2.4cm] (qbox) at (3.0, \mainY) {};

% Queue label above the box
\node[above=0.05cm of qbox, font=\small\itshape] (qlabeltxt) {Execution Queue};

% Queue items inside the box
\node[queue] at (3.0, 0.65) (q1) {$o_1$};
\node[queue] at (3.0, 0.1) (q2) {$o_2$};
\node[font=\normalsize] at (3.0, -0.35) (qdots) {$\vdots$};
\node[queue] at (3.0, -0.8) (qn) {$o_n$};

% Arrow from MAB-RRT to Queue
\draw[arrow] (mabrrt.east) -- (qbox.west);

% ===== ONLINE EXECUTION SECTION =====
% Current Object - same height as phases
\node[currobj] (currobj) at (5.3, \mainY) {Current\\Object\\$o_i$};

% Arrow from Queue to Current Object
\draw[arrow] (qbox.east) -- (currobj.west);

% Phase 1
\node[phase] (p1) at (7.5, \mainY) {Phase 1\\Grasp};
\node[output, below=0.4cm of p1] (p1out) {$q_{\text{grasp}}$,\\$T_{\text{grasp}}$};

% Phase 2
\node[phase] (p2) at (10.1, \mainY) {Phase 2\\Disassembly};
\node[output, below=0.4cm of p2] (p2out) {$\tau_{\text{dis}}$, $s_{\text{free}}$};

% Phase 3
\node[phase] (p3) at (12.7, \mainY) {Phase 3\\Goal IK};
\node[output, below=0.4cm of p3] (p3out) {$q_{\text{goal}}^{\text{weld}}$};

% Phase 4
\node[phase] (p4) at (15.3, \mainY) {Phase 4\\Regrasp};
\node[output, below=0.4cm of p4] (p4out) {$\tau_{\text{regrasp}}$};

% Phase 5
\node[phase] (p5) at (17.9, \mainY) {Phase 5\\Transport};
\node[output, below=0.4cm of p5] (p5out) {$\tau_{\text{transport}}$};

% Arrows between nodes
\draw[arrow] (currobj) -- (p1);
\draw[arrow] (p1) -- (p2);
\draw[arrow] (p2) -- (p3);
\draw[arrow] (p3) -- (p4);
\draw[arrow] (p4) -- (p5);

% Data flow arrows to outputs
\draw[dataarrow] (p1) -- (p1out);
\draw[dataarrow] (p2) -- (p2out);
\draw[dataarrow] (p3) -- (p3out);
\draw[dataarrow] (p4) -- (p4out);
\draw[dataarrow] (p5) -- (p5out);

% ===== DASHED BOXES =====
% Offline box
\node[dashedbox, fit=(assembly) (mabrrt), inner sep=12pt, 
      label={[font=\normalsize, anchor=north]north:Offline}] (offlinebox) {};

% Online Execution box  
\node[dashedbox, fit=(currobj) (p1) (p2) (p3) (p4) (p5) (p5out), inner sep=12pt,
      label={[font=\normalsize, anchor=north]north:Online Execution per Object}] (onlinebox) {};

% ===== LOOP BACK ARROW =====
\coordinate (loopstart) at ($(onlinebox.south) + (0, 0)$);
\coordinate (loopbottom) at ($(onlinebox.south) + (0, -0.7)$);
\coordinate (queueunder) at ($(qbox.south) + (0, -1.9)$);

\draw[looparrow] (loopstart) -- (loopbottom) -- (queueunder) -- (qbox.south);

% Loop label
\node[font=\normalsize\itshape, below=0.9cm of onlinebox.south, anchor=north] {repeat until queue is empty};

\end{tikzpicture}
}
\caption{Robot execution pipeline for disassembly planning. PEEL uses MAB-RRT~\cite{bayraktar2026wafr} to compute collision-free disassembly paths offline and produce an execution queue~$\sigma$ defining the extraction order. Objects are dequeued sequentially, and for each object~$o_i$ the pipeline executes five phases: (1)~grasp planning, (2)~disassembly motion following the pre-computed path until the object is free, (3)~goal inverse kinematics computation, (4)~regrasp motion to find a configuration valid at both the current (post-disassembly) and goal poses, and (5)~welded transport to the goal. The loop continues until the queue is empty.}
\label{fig:pipeline}
\end{figure*}
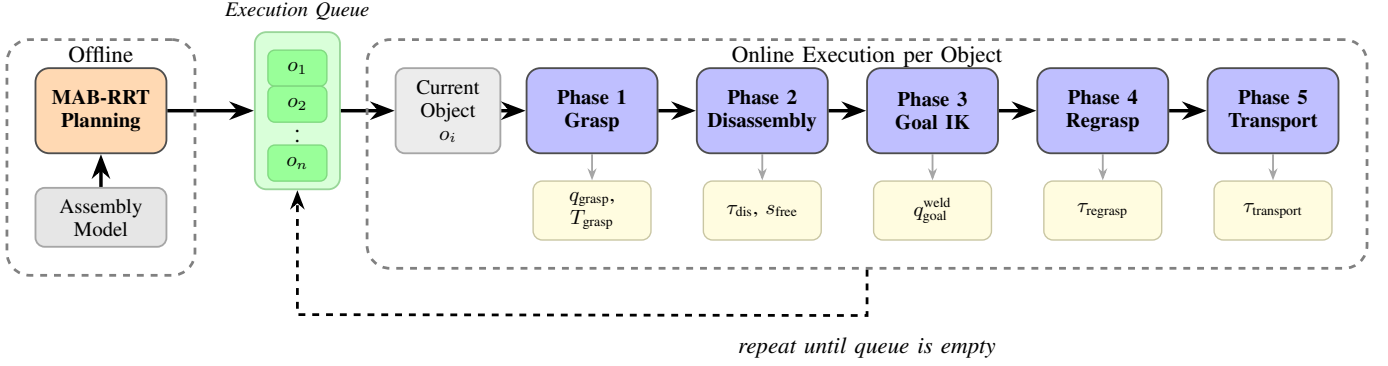

\subsection{Scale-Invariant Sampling Integration}
\label{sec:scale_sampler}\label{sec:pca_sampler}

Each single-object planner in \approachNameAbbrv is an instance of MAB-RRT~\cite{bayraktar2026wafr}, an RRT whose sampler is selected at each iteration by a multi-arm bandit. 
We summarize the components here and refer to~\cite{bayraktar2026wafr} for the full derivation. 
The \emph{scale sampler} estimates the local free scale of the configuration space around a tree node:
starting from an initial sphere radius $r_0$, it grows or shrinks the radius by fixed multiplicative factors until roughly half of a batch of sphere samples are collision-free, the point of maximum information entropy, yielding a scale $r^\star$ at which the surrounding free space is best resolved.
The \emph{PCA sampler} then exploits this scale: principal component analysis on the valid samples collected at $r^\star$ gives the dominant direction of local free space, which biases tree extensions along the narrow escape corridor, with the sampling cylinder extended by $h_{\text{ext}} = \delta \cdot r^\star$ to reach further along the passage.
The principal component is recomputed online as new valid samples accumulate. 
A sliding-window upper confidence bound (UCB) bandit arbitrates between the uniform, scale, and PCA samplers based on the reward each provides, so the planner explores when no corridor is evident and exploits the estimated direction once one is found.

\subsection{Probabilistic Completeness}

We argue that \approachNameAbbrv retains probabilistic completeness
(PC)~\cite{lavalle2006planning} on sequentially decomposable assemblies, as defined in Sec.~\ref{sec:problem_formulation}. This is true because we ensure that every object is chosen infinitely many times while using a probabilistically complete planner for which the timeout is increased to infinity. Let us formalize this:

\renewcommand*{\proofname}{Proof Sketch}

\begin{proposition}\label{prop:pc}
Let $X$ be a sequentially decomposable set of objects. Then
\approachNameAbbrv with $T_{\max} = \infty$ returns a complete disassembly sequence from $X$ with probability one.
\end{proposition}

\begin{proof}
\def\Q{\mathcal{Q}}
\def\F{\mathcal{F}}
Let $X$ be a sequentially decomposable set of $K$ objects. Since it admits an ordering, there must be at least one object $o_k \in X$ which is removable. Let us show that $o_k$ is removed with probability one by \approachNameAbbrv. First, since $o_k$ is in the queue $\Q$, we will enter the while loop in Alg.~\ref{alg:batched} and stay there until $o_k$ is removed. Since the batch size $B$ is positive and every object that fails is requeued, the queue cycles $o_k$ is selected infinitely often. In each iteration, where $o_k$ is chosen, MAB-RRT is applied to find a solution. If no solution is found, the timeout is increased and will approach infinity. Since $o_k$ is removable and MAB-RRT is probabilistically complete~\cite{bayraktar2026wafr}, it will find a solution with probability one. Once $o_k$ is removed, it is removed from the queue. Removing $o_k$ from an ordering of $X$ leaves the remaining objects so that the remainder is again sequentially decomposable.
\end{proof}

%We proceed by induction on $X$. The base case is $X$ having a single object. In that case, 
%
%$X = \emptyset$ being immediate. If
%$X$ admits a clear ordering, its first object $o_i$ is removable in $X$ by a
%path of positive clearance. Failed batches are requeued
%(line~\ref{alg:batched:requeue_all}) and recorded in $\mathcal{F}$, and $T$
%doubles only once $\mathcal{F}$ covers $\mathcal{Q}$
%(lines~\ref{alg:batched:sweepdone} and~\ref{alg:batched:double}), so no doubling
%occurs before every queued object has been tried and $o_i$ is attempted at every
%budget $2^{j}T_0$. MAB-RRT samples uniformly with positive probability at every
%iteration and so inherits probabilistic completeness from
%RRT~\cite{lavalle2006planning,bayraktar2026wafr}: writing $A_j$ for failure at
%budget $2^{j}T_0$, $\Pr(A_j) \to 0$. Never removing $o_i$ implies every $A_j$,
%an event of probability $\inf_j \Pr(A_j) = 0$, so some object is committed with
%probability one. It was removable in $X$, so \cref{cor:nodeadend} leaves a clear
%ordering on the remainder and the budget resets
%(line~\ref{alg:batched:reset}), giving the hypothesis at $|X| - 1$.
%\end{proof}
%
%Positive clearance is the usual requirement for sampling-based completeness and
%constrains the assembly rather than \approachNameAbbrv; the corridors of
%\cref{fig:page1} make it demanding, not vacuous. The cap $T_{\max}$, which buys
%termination on assemblies that admit no sequential disassembly, is the price:
%\cref{prop:pc} is recovered only as $T_{\max} \to \infty$.

\section{Robot Grounding Pipeline}\label{sec:robot_pipeline}

While \approachNameAbbrv computes collision-free paths in the object's
configuration space, executing those paths on a robot requires
bridging planning in object-space and planning in robot-space. 
To accomplish this, we introduce the robot
execution pipeline (illustrated in \cref{fig:pipeline}). This pipeline handles two
planning modes: robot-only motion, where the manipulator moves
independently while the object remains stationary, and attached motion,
where the grasped object moves rigidly with the end-effector.
To simplify grasping, we use a mobile manipulator equipped with a spherical grasping tool that
simulates stable point contacts on arbitrarily shaped surfaces.

\approachNameAbbrv produces a solution queue $\sigma$ defining the extraction order
and a set of object-space trajectories. For each object dequeued from
$\sigma$, the pipeline executes five phases.

\paragraph{Phase~I: Grasp planning}
The phase begins from the robot's current configuration $q_{\text{init}}$.
Grasp candidates are sampled near the object surface with a fixed
margin $\mu_g$ and validated through three sequential checks, any of
which may reject the candidate and trigger sampling of the next:
(i)~the candidate end-effector pose must yield a kinematically
feasible robot configuration, that is, an inverse-kinematics solution
must exist within joint limits; (ii)~the resulting configuration must
be collision-free with respect to the rest of the assembly, the
environment, and the robot itself; and (iii)~there must exist a
collision-free arm trajectory from $q_{\text{init}}$ to the grasp
configuration. The phase records the grasp
configuration $q_g$ together with the rigid transform
\begin{equation}\label{eq:tgrasp}
T_{\text{grasp}} \;=\; (T_{\text{obj}}^{(0)})^{-1} \cdot \textsc{FK}(q_g),
\end{equation}
which expresses the end-effector pose in the object's local frame and
is reused throughout Phase~II.

% Free-state detection figure -- THREE isometric mini-panels + rank strip.
% Self-contained: defines its own colors, an isometric projection, and cylinder
% helpers. Relies only on tikz + libraries arrows.meta, calc,
% decorations.pathreplacing.  Two-column-spanning figure*.
%
% (a) confined: screw in socket, r_t=1, r_r=1 (only bore axis y is free)
% (b) free state: screw cleared, r_t=3, r_r=3 -> accept/truncate
% (c) overshoot: screw in open space, path discarded toward goal region G_i
% Bottom strip: mobility rank vs. waypoint index, showing the 1->3 jump and the
% first waypoint passing the test = detected free state (aligned under panel b).
\begin{figure*}[t]
\centering
\definecolor{fsObsEdge}{RGB}{96,103,112}
\definecolor{fsObsFill}{RGB}{201,207,214}
\definecolor{fsObject}{RGB}{29,96,168}
\definecolor{fsFree}{RGB}{20,132,63}
\definecolor{fsBlock}{RGB}{201,40,40}
\definecolor{fsDiscard}{RGB}{138,142,148}
% ---- isometric projection (x lateral, y bore/axial, z up) ----
\providecommand{\fsP}[3]{({0.866*((#1)-(#2))},{0.5*((#1)+(#2))+(#3)})}
\providecommand{\fsC}[4]{%
  \pgfmathsetmacro{\fscx}{0.866*((#2)-(#3))}%
  \pgfmathsetmacro{\fscy}{0.5*((#2)+(#3))+(#4)}%
  \coordinate (#1) at (\fscx,\fscy);}
\providecommand{\fsxzc}[5]{% projected circle in x-z plane: cx cy cz R opts
  \path[#5] \fsP{#1+#4}{#2}{#3}
    \foreach \t in {6,12,...,354}{ -- \fsP{#1+#4*cos(\t)}{#2}{#3+#4*sin(\t)} }
    -- cycle;}
\providecommand{\fscap}[3]{\fsxzc{\xc}{#1}{\zc}{#2}{#3}}
% ---- projected rotation arcs: partial circles in the plane normal to an axis.
% args: centre offset along the axis, radius, start angle, sweep, draw style.
\providecommand{\fsarcx}[5]{\pgfmathsetmacro{\fsst}{(#4)/18}%
  \draw[#5] \fsP{#1}{#2*cos(#3)}{#2*sin(#3)}
    \foreach \i in {1,...,18}{ -- \fsP{#1}{#2*cos((#3)+\i*\fsst)}{#2*sin((#3)+\i*\fsst)} };}
\providecommand{\fsarcy}[5]{\pgfmathsetmacro{\fsst}{(#4)/18}%
  \draw[#5] \fsP{#2*cos(#3)}{#1}{#2*sin(#3)}
    \foreach \i in {1,...,18}{ -- \fsP{#2*cos((#3)+\i*\fsst)}{#1}{#2*sin((#3)+\i*\fsst)} };}
\providecommand{\fsarcz}[5]{\pgfmathsetmacro{\fsst}{(#4)/18}%
  \draw[#5] \fsP{#2*cos(#3)}{#2*sin(#3)}{#1}
    \foreach \i in {1,...,18}{ -- \fsP{#2*cos((#3)+\i*\fsst)}{#2*sin((#3)+\i*\fsst)}{#1} };}
% ---- clean skeleton cylinder: one flat fill + two crisp silhouette lines.
% Caps are drawn by the caller (solid near cap, dashed far rim), so the body
% carries no gradient band and no stray inner ellipse.
% args: far-end y, near-end y, fill colour, silhouette offset s (radius = s*sqrt2)
\providecommand{\fscylsurf}[5]{%
  \fill[#3]
     \fsP{\xc-#5}{#1}{\zc-#5} -- \fsP{\xc-#5}{#2}{\zc-#5}
     -- \fsP{\xc+#5}{#2}{\zc+#5} -- \fsP{\xc+#5}{#1}{\zc+#5} -- cycle;
  \draw[#4, line width=0.8pt] \fsP{\xc-#5}{#1}{\zc-#5} -- \fsP{\xc-#5}{#2}{\zc-#5};
  \draw[#4, line width=0.8pt] \fsP{\xc+#5}{#1}{\zc+#5} -- \fsP{\xc+#5}{#2}{\zc+#5};}
\providecommand{\fscylbody}[4]{\fscylsurf{#1}{#2}{#3!22!white}{#3!65!black}{#4}}
% screw head: short fat cylinder closed by a solid outer face.  The side surface
% is one tone darker than the shaft so the visible sliver reads as part of the
% head instead of a protruding lip.
\providecommand{\fsheadcyl}[2]{%
  % near-end shoulder: the annular face where the wide head meets the shaft
  \fscap{#1}{\Rh}{fill=#2!38!white, draw=#2!60!black, line width=0.5pt}%
  \fscylsurf{#1}{#1-\hL}{#2!52!white}{#2!65!black}{\sxh}
  \fscap{#1-\hL}{\Rh}{fill=#2, draw=#2!55!black, line width=0.6pt}}
\providecommand{\fsdot}[3]{%
  \ifnum#3=1 \fill[fsFree] (#1,#2) circle (0.058);
     \draw[fsFree!55!black, line width=0.3pt] (#1,#2) circle (0.058);
  \else \fill[white] (#1,#2) circle (0.058);
     \draw[fsDiscard, line width=0.6pt] (#1,#2) circle (0.052);\fi}
% rank badge: one column per object axis, so every dot names the arrow it stands
% for.  args: origin x, origin y, free flags for x, y, z.
\providecommand{\fsbadge}[5]{%
  \begin{scope}[shift={(#1,#2)}]
    \draw[badgebox] (-0.95,-0.50) rectangle (0.78,0.50);
    \node[font=\scriptsize\itshape, text=fsObsEdge, anchor=west] at (-0.90,0.31) {free?};
    \node[font=\scriptsize, anchor=west] at (-0.90,0.01)  {$r_t$};
    \node[font=\scriptsize, anchor=west] at (-0.90,-0.29) {$r_r$};
    \foreach \i/\lb/\f in {1/x/#3,2/y/#4,3/z/#5}{%
      \pgfmathsetmacro{\dx}{-0.05+0.30*(\i-1)}%
      \node[font=\scriptsize, text=fsObsEdge] at (\dx,0.31) {$\lb$};
      \fsdot{\dx}{0.01}{\f}\fsdot{\dx}{-0.29}{\f}}
  \end{scope}}
\providecommand{\fsblock}{%
  \fill[fsObsFill!70!white, edge]
     \fsP{0}{0}{\H} -- \fsP{\W}{0}{\H} -- \fsP{\W}{\D}{\H} -- \fsP{0}{\D}{\H} -- cycle;
  \fill[fsObsFill!78!black, edge]
     \fsP{0}{0}{0} -- \fsP{0}{\D}{0} -- \fsP{0}{\D}{\H} -- \fsP{0}{0}{\H} -- cycle;
  \fill[fsObsFill, edge]
     \fsP{0}{0}{0} -- \fsP{\W}{0}{0} -- \fsP{\W}{0}{\H} -- \fsP{0}{0}{\H} -- cycle;}
\resizebox{\textwidth}{!}{%
\begin{tikzpicture}[
    >=Latex,
    edge/.style       ={draw=fsObsEdge, line width=0.6pt},
    freebi/.style     ={{Latex[length=4pt]}-{Latex[length=4pt]}, fsFree, line width=1.05pt},
    blockbi/.style    ={{Latex[length=4pt]}-{Latex[length=4pt]}, fsBlock, line width=1.05pt,
                         dash pattern=on 2.2pt off 1.6pt},
    freerot/.style    ={-{Latex[length=3.6pt,width=3.6pt]}, fsFree, line width=0.9pt},
    blockrot/.style   ={-{Latex[length=3.6pt,width=3.6pt]}, fsBlock, line width=0.9pt,
                         dash pattern=on 1.9pt off 1.4pt},
    pathdash/.style   ={fsDiscard, line width=1.2pt, dash pattern=on 3.4pt off 2.6pt},
    followed/.style   ={fsObject!78!black, line width=1.4pt},
    note/.style       ={font=\footnotesize, align=center},
    ptag/.style       ={font=\small\bfseries, align=center},
    gcross/.style     ={text=fsBlock, font=\small\bfseries, inner sep=0pt},
    glab/.style       ={font=\scriptsize, text=fsObsEdge!85!black, inner sep=1.2pt},
    frame/.style      ={rounded corners=4pt, draw=fsObsEdge!55, line width=0.7pt, fill=fsObsFill!14},
    framehi/.style    ={rounded corners=4pt, draw=fsFree, line width=1.4pt, fill=fsFree!7},
    badgebox/.style   ={rounded corners=2pt, fill=white, draw=fsObsEdge, line width=0.5pt},
    axis/.style       ={fsObsEdge, line width=0.7pt},
]
\def\W{1.8}\def\D{1.6}\def\H{1.5}\def\xc{0.9}\def\zc{0.66}
\def\sxs{0.240}\def\sxh{0.335}\def\Rs{0.3394}\def\Rh{0.4738}\def\hL{0.44}
\def\PB{6.0}\def\PC{12.0}

% =============== DOF gizmo (drawn per panel via \dofgizmo) =============
% One straight double arrow (translation probe) and one curved arrow (rotation
% probe) per object axis.  Free = solid green; blocked = dashed red with a cross
% at the axis tip.  Axis letters x, y, z match the rank-badge columns.
% args: origin x, origin y, free flags for x, y, z
\newcommand{\dofgizmo}[5]{%
  \begin{scope}[shift={(#1,#2)}]
    \def\GA{0.86}\def\GC{0.64}\def\GR{0.185}\def\GT{1.05}
    % ---------- x axis (lateral, toward the viewer-right) ----------
    \ifnum#3=1 \tikzset{gax/.style=freebi, garc/.style=freerot}%
    \else      \tikzset{gax/.style=blockbi, garc/.style=blockrot}\fi
    \draw[gax] \fsP{-\GA}{0}{0} -- \fsP{\GA}{0}{0};
    \fsarcx{\GC}{\GR}{-20}{285}{garc}
    \fsC{gxp}{\GT}{0}{0}\fsC{gxn}{-\GT}{0}{0}
    \node[glab] at (gxp) {$x$};
    \ifnum#3=0 \node[gcross] at (gxn) {$\times$};\fi
    % ---------- y axis (bore / extraction axis) ----------
    \ifnum#4=1 \tikzset{gax/.style=freebi, garc/.style=freerot}%
    \else      \tikzset{gax/.style=blockbi, garc/.style=blockrot}\fi
    \draw[gax] \fsP{0}{-\GA}{0} -- \fsP{0}{\GA}{0};
    \fsarcy{-\GC}{\GR}{160}{285}{garc}
    \fsC{gyp}{0}{-\GT}{0}\fsC{gyn}{0}{\GT}{0}
    \node[glab] at (gyp) {$y$};
    \ifnum#4=0 \node[gcross] at (gyn) {$\times$};\fi
    % ---------- z axis (vertical) ----------
    \ifnum#5=1 \tikzset{gax/.style=freebi, garc/.style=freerot}%
    \else      \tikzset{gax/.style=blockbi, garc/.style=blockrot}\fi
    \draw[gax] \fsP{0}{0}{-\GA} -- \fsP{0}{0}{\GA};
    \fsarcz{-\GC}{\GR}{-160}{285}{garc}
    \fsC{gzp}{0}{0}{\GT}\fsC{gzn}{0}{0}{-\GT}
    \node[glab] at (gzp) {$z$};
    \ifnum#5=0 \node[gcross] at (gzn) {$\times$};\fi
  \end{scope}}

% ================= progression arrow spanning the panels ==============
\draw[-{Latex[length=5pt]}, line width=1.1pt, fsObsEdge] (-0.7,4.80) -- (6.9,4.80);
\node[note] at (3.1,5.03) {extraction along bore axis $y$};
\draw[-{Latex[length=5pt]}, pathdash] (6.9,4.80) -- (15.6,4.80);
\node[note, text=fsDiscard] at (11.4,5.03)
   {path truncated at free state $\Rightarrow$ overshoot discarded};
\draw[fsFree, line width=1.0pt] (6.9,4.57) -- (6.9,5.03);

% =====================================================================
%  PANEL (a) CONFINED
% =====================================================================
\begin{scope}
  \draw[frame] (-1.7,-1.4) rectangle (3.95,4.30);
  \node[note, anchor=west] at (-1.55,3.98) {screw in socket};
  \fsblock
  \fscap{0}{0.42}{fill=fsObsFill!55!black, draw=fsObsEdge, line width=0.5pt}
  \fscap{0}{\Rs}{fill=fsObject!24!white, draw=fsObject!65!black, line width=0.7pt}
  \fscylbody{0}{-0.53}{fsObject}{\sxs}
  \fsheadcyl{-0.53}{fsObject}
  \fsC{ha}{\xc}{-0.97}{\zc}
  \node[note, anchor=north] at ($(ha)+(0.05,-0.60)$) {$T_{obj}^{(0)}$};
  \dofgizmo{2.70}{1.95}{0}{1}{0}
  \fsbadge{2.78}{3.74}{0}{1}{0}
  \node[ptag] at (0.95,-1.05) {(a) confined};
\end{scope}

% =====================================================================
%  PANEL (b) FREE STATE  (highlighted)
% =====================================================================
\begin{scope}[shift={(\PB,0)}]
  \draw[framehi] (-1.7,-1.4) rectangle (3.95,4.30);
  \node[note, text=fsFree, font=\small\bfseries, anchor=west] at (-1.55,3.98) {screw clears socket};
  \fsblock
  \fscap{0}{0.42}{fill=fsObsFill!42!black, draw=fsObsEdge, line width=0.5pt}
  \fscap{-0.85}{\Rs}{draw=fsObject!60!black, line width=0.5pt, dash pattern=on 1.6pt off 1.4pt}
  \fscylbody{-0.85}{-1.90}{fsObject}{\sxs}
  \fsheadcyl{-1.90}{fsObject}
  \fsC{hb}{\xc}{-2.34}{\zc}
  \node[note, anchor=north] at ($(hb)+(0.05,-0.50)$) {$T_{obj}^{(k^\star)}$};
  \dofgizmo{2.70}{1.95}{1}{1}{1}
  \fsbadge{2.78}{3.74}{1}{1}{1}
  \node[note, text=fsFree, font=\small\bfseries] at (0.30,-0.70) {accept \& truncate here};
  \node[ptag] at (0.95,-1.05) {(b) free state};
\end{scope}

% =====================================================================
%  PANEL (c) OVERSHOOT  (sparse, discarded);  G_i moved up & away
% =====================================================================
\begin{scope}[shift={(\PC,0)}]
  \draw[frame] (-1.7,-1.4) rectangle (4.4,4.30);
  \node[note, text=fsDiscard, anchor=west] at (-1.55,3.98) {overshoot (discarded)};
  \fsblock
  \fscap{0}{0.42}{fill=fsObsFill!42!black, draw=fsObsEdge, line width=0.5pt}
  % discarded path: mouth -> ghost, ghost -> G_i (up, away from the screw)
  \draw[pathdash] \fsP{\xc}{0}{\zc} -- \fsP{\xc}{-1.4}{\zc};
  \draw[pathdash] \fsP{\xc}{-2.7}{\zc} -- (2.65,2.25);
  \fscap{-1.4}{\Rs}{draw=fsDiscard!60!black, line width=0.5pt, dash pattern=on 1.6pt off 1.4pt}
  \fscylbody{-1.4}{-2.4}{fsDiscard}{\sxs}
  \fsheadcyl{-2.4}{fsDiscard}
  \node[note, text=fsDiscard, anchor=north] at (1.90,-0.32) {$T_{obj}^{(N)}$};
  % path continues off toward the distant goal region (no box, keep it sparse)
  \node[note, text=fsDiscard, font=\scriptsize, anchor=south] at (2.65,2.32) {toward $G_i$};
  \node[ptag] at (0.55,-1.05) {(c) overshoot};
\end{scope}

% =====================================================================
%  RANK STRIP : mobility rank along the extraction path
% =====================================================================
% mapping: x(k)=0.9+0.98 k ; y(r)=-3.55+0.42 r ; detection at k=6 (~panel b)
\def\ry{-3.95}                       % r=0 baseline
\draw[rounded corners=3pt, draw=fsObsEdge!55, line width=0.7pt, fill=fsObsFill!10]
   (-1.7,-4.55) rectangle (16.5,-1.75);
% axes
\draw[axis, -{Latex[length=4pt]}] (-0.55,\ry) -- (15.4,\ry)
   node[note, anchor=north east, yshift=-1pt] {waypoint index $k$};
\draw[axis, -{Latex[length=4pt]}] (-0.55,\ry) -- (-0.55,{\ry+3*0.42+0.28});
\foreach \r in {0,1,2,3}{ \draw[axis] (-0.63,{\ry+\r*0.42}) -- (-0.55,{\ry+\r*0.42});
   \node[font=\scriptsize, anchor=east] at (-0.66,{\ry+\r*0.42}) {\r}; }
\node[font=\scriptsize\bfseries, anchor=south, rotate=90] at (-1.15,{\ry+1.5*0.42}) {rank};
% thresholds
\draw[fsObsEdge, dash pattern=on 3pt off 2pt] (-0.55,{\ry+3*0.42}) -- (13.6,{\ry+3*0.42})
   node[font=\scriptsize, anchor=west, text=fsObsEdge] {$\rho_t=3$};
\draw[fsObsEdge!70, dash pattern=on 2pt off 2pt] (-0.55,{\ry+2*0.42}) -- (13.6,{\ry+2*0.42})
   node[font=\scriptsize, anchor=west, text=fsObsEdge!70] {$\rho_r=2$};
% r_r step (thin, secondary): 1 until k=6, then 3  (dashed after truncation)
\draw[fsObject!62!white, line width=0.9pt]
   (0.9,{\ry+1*0.42-0.10}) -- (6.78,{\ry+1*0.42-0.10}) -- (6.78,{\ry+3*0.42-0.10});
\draw[fsDiscard, line width=0.9pt, dash pattern=on 2.4pt off 2pt]
   (6.78,{\ry+3*0.42-0.10}) -- (13.4,{\ry+3*0.42-0.10});
\node[font=\scriptsize, text=fsObject!75!white, anchor=west] at (3.0,{\ry+1*0.42-0.28}) {$r_r$};
% r_t step (bold): 1 until k=6, jump to 3 ; followed solid then dashed overshoot
\draw[followed] (0.9,{\ry+1*0.42}) -- (6.78,{\ry+1*0.42});
\draw[fsFree, line width=1.4pt] (6.78,{\ry+1*0.42}) -- (6.78,{\ry+3*0.42});
\draw[pathdash] (6.78,{\ry+3*0.42}) -- (13.4,{\ry+3*0.42});
\node[font=\scriptsize\bfseries, text=fsObject!78!black, anchor=east] at (6.4,{\ry+1*0.42+0.16}) {$r_t$};
% detection marker + tie line up to panel (b)
\fill[fsFree] (6.78,{\ry+3*0.42}) circle (0.09);
\draw[fsFree, line width=0.9pt, dash pattern=on 2.5pt off 2pt] (6.78,{\ry+3*0.42}) -- (6.78,-1.55);
% strip title drawn last, with an opaque chip so the tie-line passes behind it
\node[note, anchor=west, fill=fsObsFill!10, inner sep=2pt] at (-1.55,-1.98)
   {\small mobility rank tested at each waypoint; the \textbf{first} waypoint with
    $r_t\ge\rho_t$ \emph{and} $r_r\ge\rho_r$ is the free state};

% =================== shared acceptance chip + legend =================
\draw[rounded corners=3pt, fill=white, draw=fsObsEdge, line width=0.6pt]
   (-1.7,-6.15) rectangle (16.5,-4.8);
\node[note, anchor=west] at (-1.5,-5.15)
   {\textbf{accept the free state when} $r_t\ge\rho_t$ \textbf{and} $r_r\ge\rho_r$
    \;\; (example $(\rho_t,\rho_r)=(3,2)$)};
\draw[freebi] (-1.5,-5.75) -- (-1.0,-5.75);
\draw[freerot] (-0.84,-5.88) arc[start angle=-70, end angle=200, radius=0.14];
\node[note, anchor=west] at (-0.62,-5.75)
   {free axis ($\pm\varepsilon_t$, $\pm\varepsilon_r$)};
\draw[blockbi] (2.5,-5.75) -- (3.0,-5.75);
\draw[blockrot] (3.16,-5.88) arc[start angle=-70, end angle=200, radius=0.14];
\node[gcross] at (3.46,-5.75) {$\times$};
\node[note, anchor=west] at (3.64,-5.75) {blocked axis};
\fsdot{5.60}{-5.69}{1}\fsdot{5.82}{-5.69}{0}
\node[note, anchor=west] at (5.98,-5.75) {badge dot: free / blocked};
\draw[followed] (9.8,-5.75) -- (10.3,-5.75);
\node[note, anchor=west] at (10.4,-5.75) {followed path};
\draw[pathdash] (12.5,-5.75) -- (13.0,-5.75);
\node[note, anchor=west] at (13.1,-5.75) {discarded overshoot};

\end{tikzpicture}%
}
\caption{Free-state detection while withdrawing a screw from a socket, shown as
three isometric stages (mobility rank is inherently three-dimensional, hence the
3D view) with, below, the mobility rank evaluated along the whole extraction
path. At each waypoint the local mobility is probed with small
$\pm\varepsilon_t$ translations along and $\pm\varepsilon_r$ rotations about the
three object axes $x$, $y$, $z$, where $y$ is the bore axis. The translational
(rotational) rank $r_t$ ($r_r$) is the dimension spanned by the free directions,
which here are axis aligned, and the rank badge repeats it with one column per
axis, so every dot names the arrow it stands for. In this example the same axes
are free in translation and in rotation, so a single colour per axis carries
both. \textbf{(a)} While the screw is still in the socket
only the axial slide and the axial spin about $y$ are free ($r_t=1$, $r_r=1$);
both lateral axes $x$ and $z$ are blocked by the socket wall. \textbf{(b)} The
instant the screw clears the socket all six degrees of freedom open up ($r_t=3$,
$r_r=3$), so this waypoint $T_{obj}^{(k^\star)}$ is accepted as the free state
once $r_t\ge\rho_t$ and $r_r\ge\rho_r$ (here $(\rho_t,\rho_r)=(3,2)$) and the
path is truncated there. \textbf{(c)} The raw path continues in open space toward
the distant goal region $G_i$, so this overshoot $T_{obj}^{(N)}$ is discarded
and replanned as free-space transport. The bottom strip makes the mechanism
explicit: the rank stays at $1$ while the screw is in the socket and jumps to
$3$ the instant it clears; the first waypoint crossing the thresholds (green
marker, aligned under panel~(b)) is the detected free state.}
\label{fig:free_state}
\end{figure*}
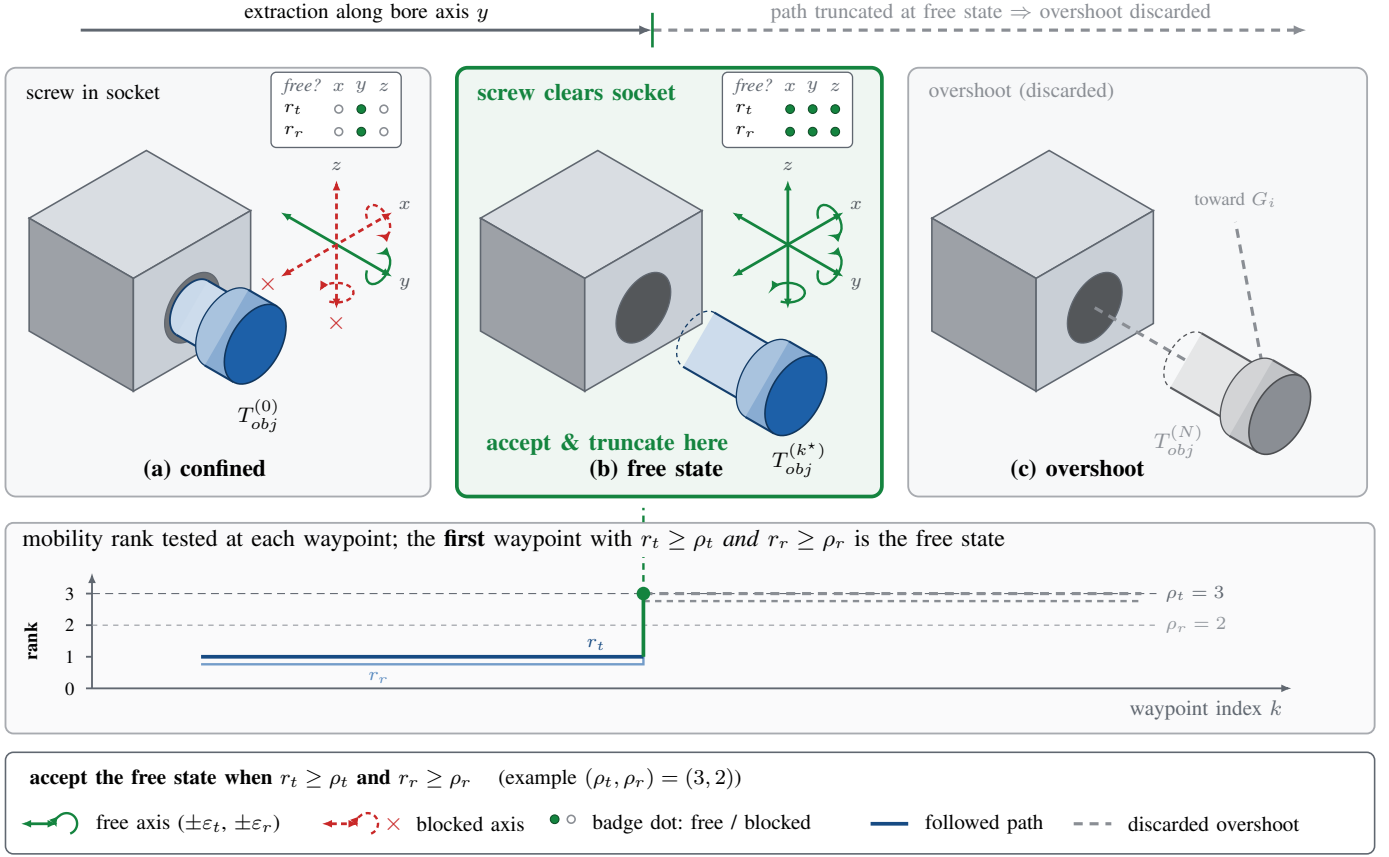
\paragraph{Phase~II: Disassembly motion}
MAB-RRT plans the object's motion as a tree growing from the start
configuration into free space, but the path it returns does not
explicitly indicate where the narrow-passage portion of the extraction
ends and where the object becomes effectively unconstrained. We
require this boundary because the goal of Phase~II is the
narrow-passage portion only: once the object can move freely, the
remaining transport to the goal pose can be replanned by a standard
RRT in Phase~V without the directional bias of MAB-RRT, often along a
much shorter path than the one emitted by the search tree.

A first instinct is to cut the path at the first or last waypoint of a
particular sampling arm, but this is unreliable. By the time the object is free
the search has long converged on the PCA arm and keeps extending toward the
distant goal $G_i$, so the last waypoint typically sits far outside the assembly,
in mid-air (\cref{fig:free_state}), and the uniform arm can place a valid sample
almost anywhere. Neither arm gives a reliable cut between the narrow-passage and
free-space portions of the path.

% A first instinct is to derive the boundary from the path itself, for
% example by truncating it at the first or last waypoint produced by a
% particular sampling arm. This heuristic is unreliable in practice for
% several reasons. By the time the object is geometrically free the MAB
% has long since converged on the PCA arm and continues to extend the
% tree in the principal escape direction toward the goal region $G_i$; the search therefore keeps
% adding waypoints far past the point at which the object was already
% unconstrained, so the last PCA waypoint typically sits well outside
% the assembly, effectively in mid-air. Truncating there would commit
% the manipulator to dragging the object out to that arbitrary
% location.
% The uniform arm fares no better: by construction its samples can land
% anywhere in the workspace, so the last accepted uniform waypoint may
% be similarly suspended in free space far from the assembly, while
% nothing prevents a valid uniform sample from also coming from a
% configuration where the object is still partially confined by the
% surrounding parts. Arguing from the first or last occurrence of a
% particular arm therefore provides no reliable cut between the
% narrow-passage and the free-space portions of the path.

We instead detect the free state by probing the object's local
\emph{mobility} at each waypoint. Around the current object pose
$T_{\text{obj}}^{(k)}$, we perturb the object by $\pm\epsilon_t$ along each
translational axis and by $\pm\epsilon_r$ about each rotational axis,
collision-checking every perturbation against the rest of the assembly.
Collecting collision-free directions, the translational mobility rank
$r_t \in \{0,\dots,3\}$ is the dimension of the subspace they span, computed as
the numerical rank of that set of directions; the rotational mobility rank
$r_r \in \{0,\dots,3\}$ is defined analogously. The waypoint is accepted as a
free state when $r_t \geq \rho_t$ and $r_r \geq \rho_r$. With $\rho_t = 3$ and
$\rho_r = 2$, the criterion requires the collision-free translational directions
to span all three dimensions and the rotational ones to span at least two. This is a conservative geometric definition of
\emph{escaped}: the object is unobstructed by the surrounding parts in a
local neighborhood, and the manipulator can therefore reorient and
translate it with a standard motion planner from this point onward.

The trajectory is followed by mapping each accepted waypoint to a
target end-effector pose using
\begin{equation}\label{eq:ee_target}
T_{\text{ee}}^{\text{target}} \;=\; T_{\text{obj}}^{(k)} \cdot T_{\text{grasp}},
\end{equation}
and solving inverse kinematics with the previous configuration as a seed
to encourage smooth joint motion. Following a precomputed object
trajectory exposes the manipulator to several failure modes that do
not arise during free-space planning. The joint configuration may
approach a joint limit as the object rotates relative to the base;
the end-effector may approach the boundary of the reachable
workspace; the arm may pass through a kinematic singularity at which
small target motions demand large joint motions; and the arm itself
may collide with another part of the assembly even when the object
remains collision-free along its trajectory. Each of these manifests
as an IK or collision failure at an otherwise valid object waypoint.
We respond with an intermediate \emph{regrasp}: the robot releases
the object in place, a new grasp candidate is sampled and validated
at the current object pose, a robot-only path is planned to the new
grasp configuration, and trajectory following resumes
with an updated grasp transform $T_{\text{grasp}}$. Regrasping
changes which sub-region of the configuration space the manipulator
must traverse to keep the object on its trajectory: a different grasp
can sidestep an approaching joint limit, leave the workspace
boundary, escape a singular pose, or route the arm clear of an
obstacle that the previous grasp pinned it against. As a result,
extractions whose total motion exceeds the kinematic, workspace, or
local-geometry reach of any single grasp can still complete in a
single Phase~II execution.

\paragraph{Phase~III: Goal IK computation}
The grasp transform $T_{\text{grasp}}$ accepted in Phase~I was
selected jointly with the start configuration of the object and the
geometry of the surrounding parts. Once the object is free and the
remaining task is to place it at its goal pose $T_{\text{obj}}^{\text{goal}}$,
the same transform is unlikely to remain feasible: the workspace and
the obstacles around the goal are different from those around the
start, and the end-effector frame that was suitable for threading the
object through a narrow passage may now intersect the table or fall
outside the manipulator's reach. We therefore search for a fresh
grasp at the goal pose, with the additional constraint that it must
also be reachable from the object's current free-state pose so that
the robot can switch to it before transport.

In this phase, we sample candidate end-effector poses $p_{\text{ee}}$ near
the object surface (with margin $\mu_g$, as in Phase~I) and treats
each candidate as a tentative goal grasp. For a candidate to be
accepted, two inverse-kinematics queries must succeed simultaneously.
First, the candidate must yield a collision-free configuration
$q_{\text{goal}}$ that places the end-effector at
$T_{\text{obj}}^{\text{goal}} \cdot p_{\text{ee}}$, ensuring the
object can be physically deposited at its target with the new grasp.
Second, the same grasp transform
$T_{\text{grasp}}' = (T_{\text{obj}}^{\text{goal}})^{-1} \cdot
\textsc{FK}(q_{\text{goal}})$ must yield a collision-free
configuration $q_{\text{br}}$ when applied at the object's current
free-state pose $T_{\text{obj}}^{(*)}$, providing the configuration
the robot needs to reach before it can re-attach to the object with
$T_{\text{grasp}}'$ and transport it. Candidates that fail either
query are discarded, and the loop resamples up to $N_g'$ times.

Requiring a single grasp transform to be simultaneously collision-free and within joint limits at both the free-state pose $T_{\text{obj}}^{(*)}$ and the goal pose $T_{\text{obj}}^{\text{goal}}$ is a nontrivial constraint. The two poses can demand geometrically incompatible approach directions, for example when a screw threaded out from above must be set down from the side, so that for some object and goal pairs no shared grasp exists at all. Phase~III searches explicitly for such a grasp, and its failure after $N_g'$ samples signals a genuine infeasibility rather than a mere sampling shortfall.

The
resulting triple $(T_{\text{grasp}}',\, q_{\text{goal}},\, q_{\text{br}})$
defines the boundary configurations of the next two phases. This
sequential approach (search at the goal first, then verify at the
free state) is enabled by the purely geometric formulation
introduced in \cref{sec:problem_formulation}: in the absence of
gravity and contact forces, the object can be presumed to remain at
$T_{\text{obj}}^{(*)}$ while the robot relinquishes and re-attaches
its grasp.

\paragraph{Phase~IV: Bridge motion}
Phase~IV realizes the regrasp implied by Phase~III. We plan
a robot-only path from the Phase~II final configuration to the bridge
configuration $q_{\text{br}}$, during which the object is presumed to
remain stationary at $T_{\text{obj}}^{(*)}$. 
At the end of the phase
the end-effector reaches $T_{\text{obj}}^{(*)} \cdot T_{\text{grasp}}'$,
that is, the new grasp transform applied at the object's current
pose, and the robot is in position to re-attach to the object for
transport.

If the planner fails to find such a path the phase reports failure and
recovery requires re-invoking Phase~III to draw a fresh goal grasp,
since the $q_{\text{br}}$ accepted by Phase~III was only tested for being
collision-free at $T_{\text{obj}}^{(*)}$ and not for reachability
from the Phase~II end configuration. 
%A cleaner architectural
%alternative would fold the reachability test into Phase~III's
%acceptance criterion by extending the per-candidate check with an
%RRTConnect query from the Phase~II end configuration to
%$q_{\text{br}}$, rejecting candidates for which no such path exists.
%Phase~III would then only return triples for which Phase~IV is
%guaranteed feasible, eliminating the inter-phase failure path at the
%cost of one additional motion-planning call per accepted candidate;
%we leave this consolidation as a follow-on improvement.

\paragraph{Phase~V: Transport}
The object is rigidly welded to the end-effector via the new grasp
$T_{\text{grasp}}'$ and transported to its goal pose by a
plan from $q_{\text{br}}$ to $q_{\text{goal}}$. During this phase the
object moves rigidly with the end-effector under the welding
constraint, so collision checking is performed on the full
robot-and-object assemblage rather than on the manipulator alone.
The pipeline then dequeues the next object and repeats until
$\sigma$ is empty.
\section{Evaluation}

We evaluate \approachNameAbbrv, MAB-RRT, and the robot execution pipeline on two complementary benchmark sets. First, we evaluate MAB-RRT on single-part disassembly using the
Automate dataset~\cite{tang2024automate}. This dataset contains over 100
mechanical assemblies. However, since it includes initial collisions while our
method requires strict separation, we filter to the 76 collision-free
models to establish a common benchmark. Each model is tested under 10
random rotations, yielding 760 trials. The benchmark is described in Sec.~\ref{sec:single_part_benchmarks}. Second, for multi-part
disassembly, we evaluate on 4 assemblies with 10--17 removable parts
also taken from the Automate dataset. The benchmark is described in Sec.~\ref{sec:multi_part_benchmarks}. Finally, we demonstrate the robot execution pipeline by using the sequences from \approachNameAbbrv on the four assemblies as input. This is showcased in Sec.~\ref{sec:robot_execution_demonstration}.

\paragraph{Hardware and Implementation}

All experiments were conducted on Ubuntu 20.04.6 LTS with an Intel Core
i7-5820K CPU (3.30\,GHz, 12 cores) and 32\,GB RAM. MAB-RRT~\cite{bayraktar2026wafr} is
implemented in C++ as an extension of
OMPL~\cite{sucan2012the-open-motion-planning-library}. All experiments,
covering single-object benchmarks and the robot execution pipeline,
run through Robowflex~\cite{kingston2022robowflex}, a C++ wrapper
around MoveIt that integrates OMPL planners and uses
DARTSim~\cite{lee2018dart} for collision checking.
Single-object benchmarks run single-threaded; multi-part batching
(\cref{sec:parallel_batched}) uses 10 concurrent processes per batch.
AssembleThemAll (ATA)~\cite{tian2022assemble} baselines were executed
from the authors'
repository\footnote{\url{https://github.com/yunshengtian/Assemble-Them-All}}
without modification. 

%%%%%%%%%%%%%%%%%%%%%%%%%%%%%%%%%%%%%%%%%%%%%%%%
\subsection{Single-Object Benchmarks\label{sec:single_part_benchmarks}}
%%%%%%%%%%%%%%%%%%%%%%%%%%%%%%%%%%%%%%%%%%%%%%%%
\begin{table}[t]
\centering
\caption{Combined planner comparison across 76 collision-free Automate
  assemblies (760 trials, 10 random rotations each).}
\label{tab:combined_comparison}
\begin{tabular}{lccccc}
\toprule
Planner & Success Rate & Mean & Median & Std Dev & Runs \\
 & (\%) & (s) & (s) & (s) & \\
\midrule
\textbf{MAB-RRT} & \textbf{100.0} & \textbf{4.31} & \textbf{3.20} & \textbf{3.79} & 760 \\
BFS & 53.9 & 44.59 & 38.23 & 31.26 & 760 \\
RRT (bridge) & 47.5 & 22.15 & 7.63 & 29.93 & 760 \\
RRT (gaussian) & 47.1 & 18.63 & 6.37 & 27.20 & 760 \\
RRT (obstacle) & 45.1 & 20.27 & 6.99 & 27.92 & 760 \\
TRRT & 23.0 & 54.72 & 48.05 & 40.06 & 760 \\
MateVec-TRRT & 22.2 & 56.68 & 49.71 & 39.59 & 760 \\
BK-RRT & 21.8 & 46.31 & 43.26 & 32.76 & 760 \\
RRT (uniform) & 11.4 & 57.52 & 49.45 & 49.13 & 760 \\
\bottomrule
\end{tabular}
\end{table}

\begin{figure}[t]
    \centering
    \includegraphics[width=\linewidth]{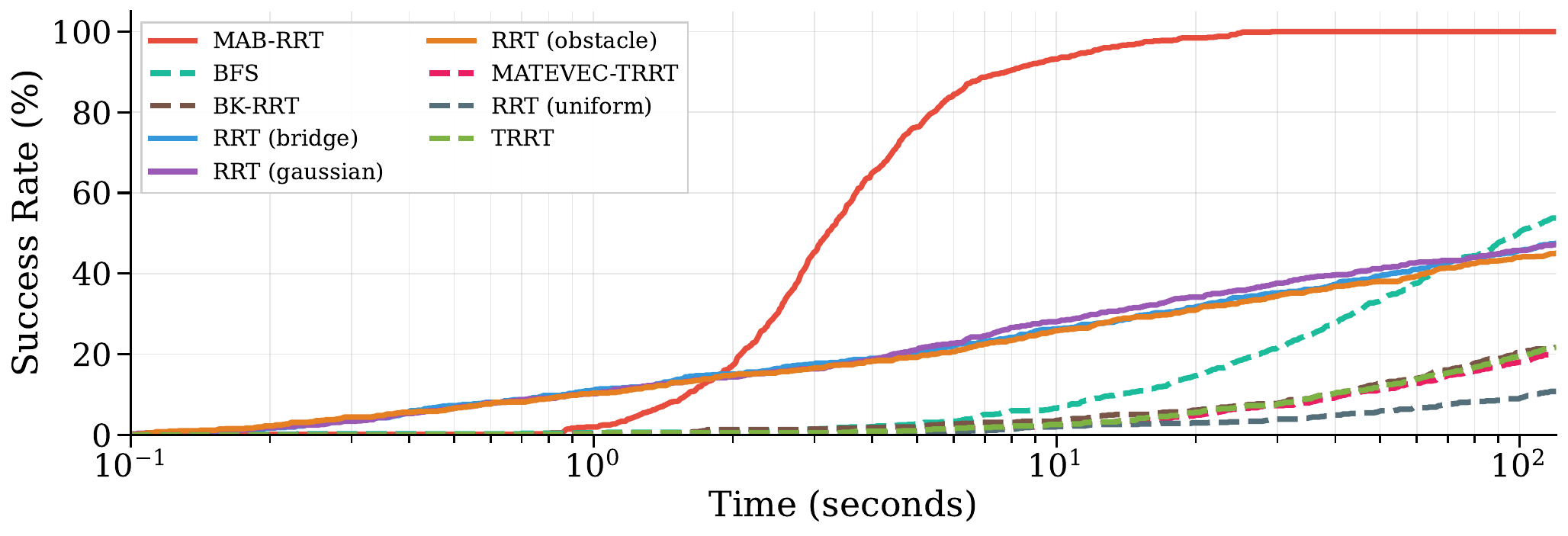}
    \caption{Success rate progression across 76 benchmark assemblies
      (760 trials, 10 random rotations per assembly). Each curve shows
      the fraction of the 760 trials that completed within elapsed
      wall-clock time $t$ (denominator fixed at 760). Solid lines
      represent MAB-RRT and RRT with informed sampling strategies
      (bridge, Gaussian, obstacle) from OMPL; dashed lines represent
      ATA baseline planners (BFS, BK-RRT, MateVec-TRRT, RRT uniform,
      TRRT). MAB-RRT achieves 100\% success within seconds,
      significantly outperforming all baselines.}
    \label{fig:success_progression}
\end{figure}

\begin{figure}[t]
    \centering
    \includegraphics[width=\linewidth]{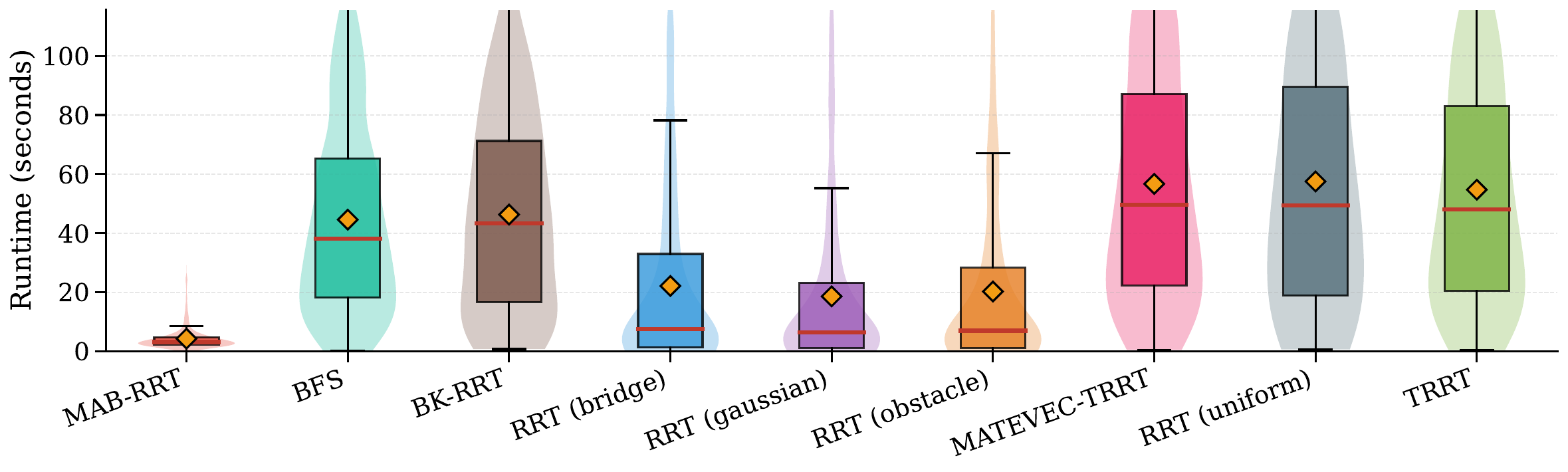}
    \caption{Runtime distribution for successful planning runs across
      all benchmark assemblies. Violin plots show the probability
      density of solution times, with overlaid box plots displaying
      quartile statistics. MAB-RRT achieves the fastest median and mean
      solution times with substantially lower variance than all baselines.}
    \label{fig:runtime_violin}
\end{figure}

We compare MAB-RRT against two sets of planners. The first set
integrates classical sampling strategies into
RRT~\cite{Kuffner2000}: bridge
sampling~\cite{Hsu2003BridgeTest}, Gaussian
sampling~\cite{Boor1999GaussianSampling}, and obstacle-based
sampling~\cite{Amato1998ObstacleBased}. The second set consists of
modern strategies from the ATA~\cite{tian2022assemble} framework: mating vectors
(MateVec-TRRT)~\cite{Ebinger2018MateVecTRRT}, behavioral kinodynamic
RRT (BK-RRT)~\cite{zickler2009efficient,tian2022assemble},
disassembly breadth-first search (BFS)~\cite{tian2022assemble},
Transition-based RRT (TRRT)~\cite{tian2022assemble}, and plain uniform
RRT. For the classical strategies, we use the default parameters as
specified in
OMPL~\cite{sucan2012the-open-motion-planning-library,moll2015benchmarking-motion-planning-algorithms}.
For the ATA strategies, we use the default parameters from the ATA
software package.

\paragraph{Success Criteria and Collision Resolution}

MAB-RRT requires an exact collision-free path from the initial
configuration to a free-space goal. We set OMPL's
\texttt{Longest\-Valid\-Segment\-Fraction} to $10^{-4}$, forcing a
collision check every $0.01\%$ of the workspace extent. ATA success is defined as
termination in their planner-defined disassembled state with their
standard step size of $10^{-2}$.

\paragraph{Results}
To evaluate all planners on the 76 collision-free assemblies from the
Automate dataset~\cite{tang2024automate} (760 trials total), we use a per-trial timeout of
$120$\,s. 
\cref{tab:combined_comparison} summarizes the results,
\cref{fig:success_progression} shows success rate progression over time,
and \cref{fig:runtime_violin} shows runtime distributions for
successful runs.

MAB-RRT achieves a 100\% success rate with a mean runtime of 4.31\,s
and a median of 3.20\,s. The next best planners are BFS (53.9\%),
RRT with bridge sampling (47.5\%), RRT with Gaussian sampling
(47.1\%), and RRT with obstacle sampling (45.1\%). The ATA-specific
planners, MateVec-TRRT (22.2\%), TRRT
(23.0\%), and BK-RRT (21.8\%), all remain below 25\% success rate,
while plain uniform RRT solves only 11.4\% of trials. The success rate
progression in \cref{fig:success_progression} shows that MAB-RRT
reaches full success within the first few seconds, while all baselines
plateau well below 60\%. The curve is computed by counting, at each
elapsed wall-clock time $t$, the fraction of the 760 trials (76
assemblies $\times$ 10 random rotations) that have completed within
$t$. A trial completes when the planner returns a collision-free
extraction path or reaches the per-trial wall-clock budget. The
denominator is fixed at 760 throughout, so a curve plateauing at
$0.6$ corresponds to 456 successful trials out of 760.

%%%%%%%%%%%%%%%%%%%%%%%%%%%%%%%%%%%%%%%%%%%%%%%%
\subsection{Multi-Part Disassembly Benchmark\label{sec:multi_part_benchmarks}}
%%%%%%%%%%%%%%%%%%%%%%%%%%%%%%%%%%%%%%%%%%%%%%%%
\begin{figure}
    \centering
    
    % ================== ADJUSTABLE WIDTHS ==================
    % Change these values (should sum close to 1.0)
    \newcommand{\assemblyleftfrac}{0.32}   % Left = assembled
    \newcommand{\assemblyrightfrac}{0.65}  % Right = exploded
    
    % ================== MICROSCOPE ==================
    \begin{subfigure}[c]{\assemblyleftfrac\linewidth}
        \centering
        \includegraphics[width=\linewidth,height=3.5cm,keepaspectratio]{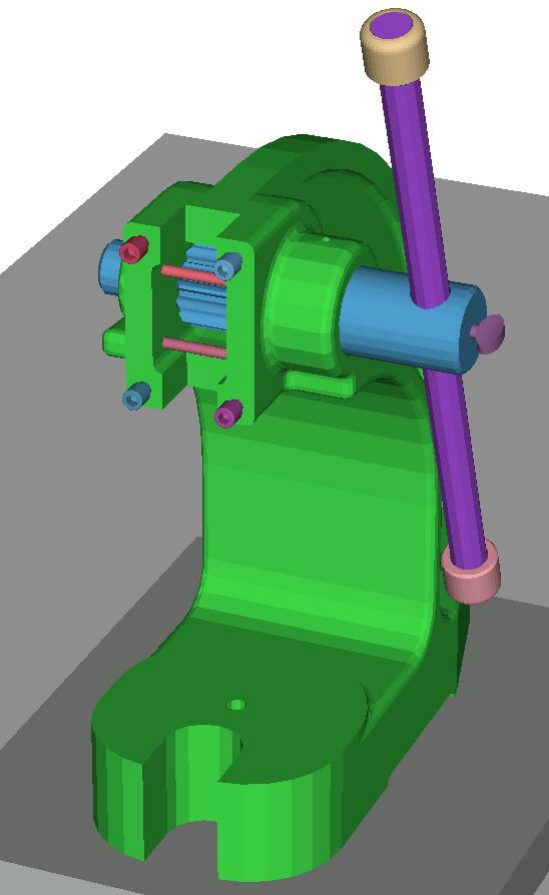}
        \caption*{Microscope (assembled)}
    \end{subfigure}
    \hfill
    \begin{subfigure}[c]{\assemblyrightfrac\linewidth}
        \centering
        \includegraphics[width=\linewidth,height=3.5cm,keepaspectratio]{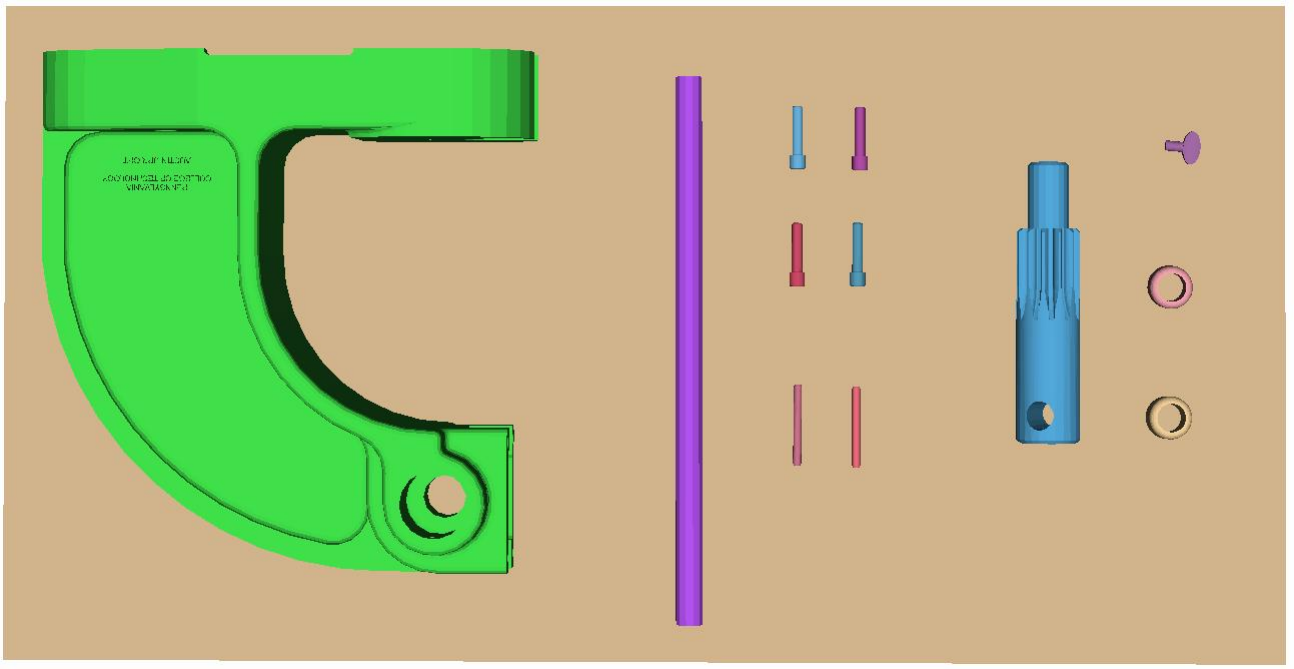}
        \caption*{Microscope (exploded): 12 parts}
    \end{subfigure}
    % ================== DISC BRAKE ==================
    \begin{subfigure}[c]{\assemblyleftfrac\linewidth}
        \centering
        \includegraphics[width=\linewidth,height=3.5cm,keepaspectratio]{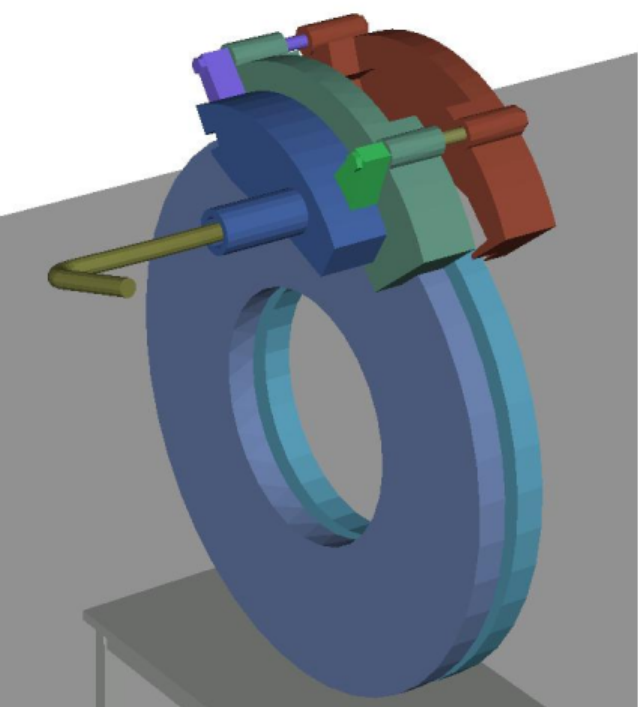}
        \caption*{Disc Brake (assembled)}
    \end{subfigure}
    \hfill
    \begin{subfigure}[c]{\assemblyrightfrac\linewidth}
        \centering
        \includegraphics[width=\linewidth,height=3.5cm,keepaspectratio]{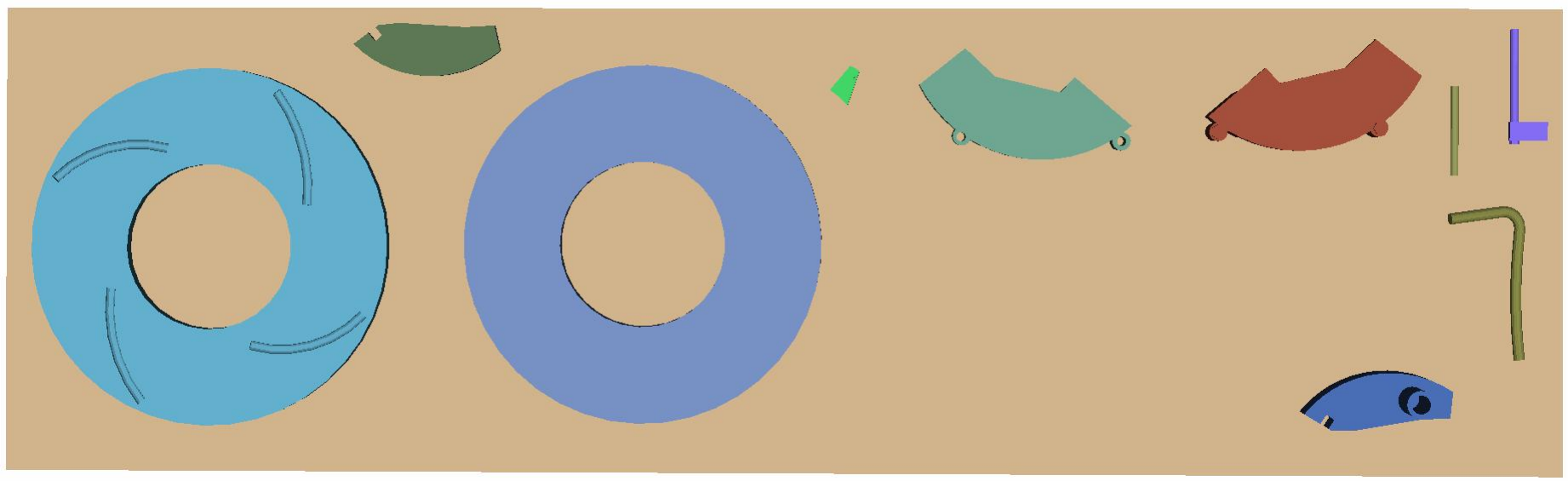}
        \caption*{Disc Brake (exploded): 10 parts}
    \end{subfigure}
    % ================== COUPLING BLOCK ==================
    \begin{subfigure}[c]{\assemblyleftfrac\linewidth}
        \centering
        \includegraphics[width=\linewidth,height=3.5cm,keepaspectratio]{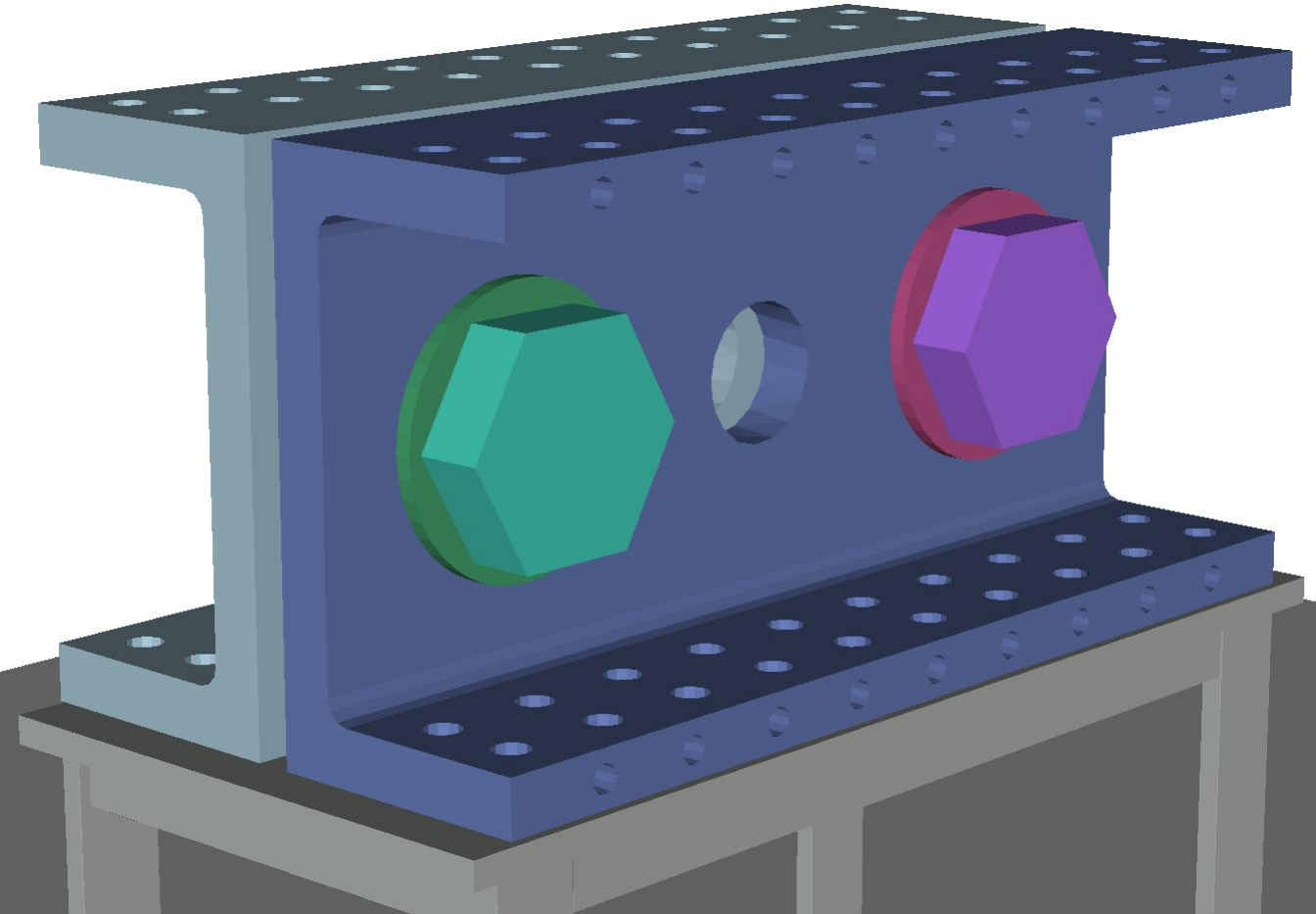}
        \caption*{Coupling Block (assembled)}
    \end{subfigure}
    \hfill
    \begin{subfigure}[c]{\assemblyrightfrac\linewidth}
        \centering
        \includegraphics[width=\linewidth,height=3.5cm,keepaspectratio]{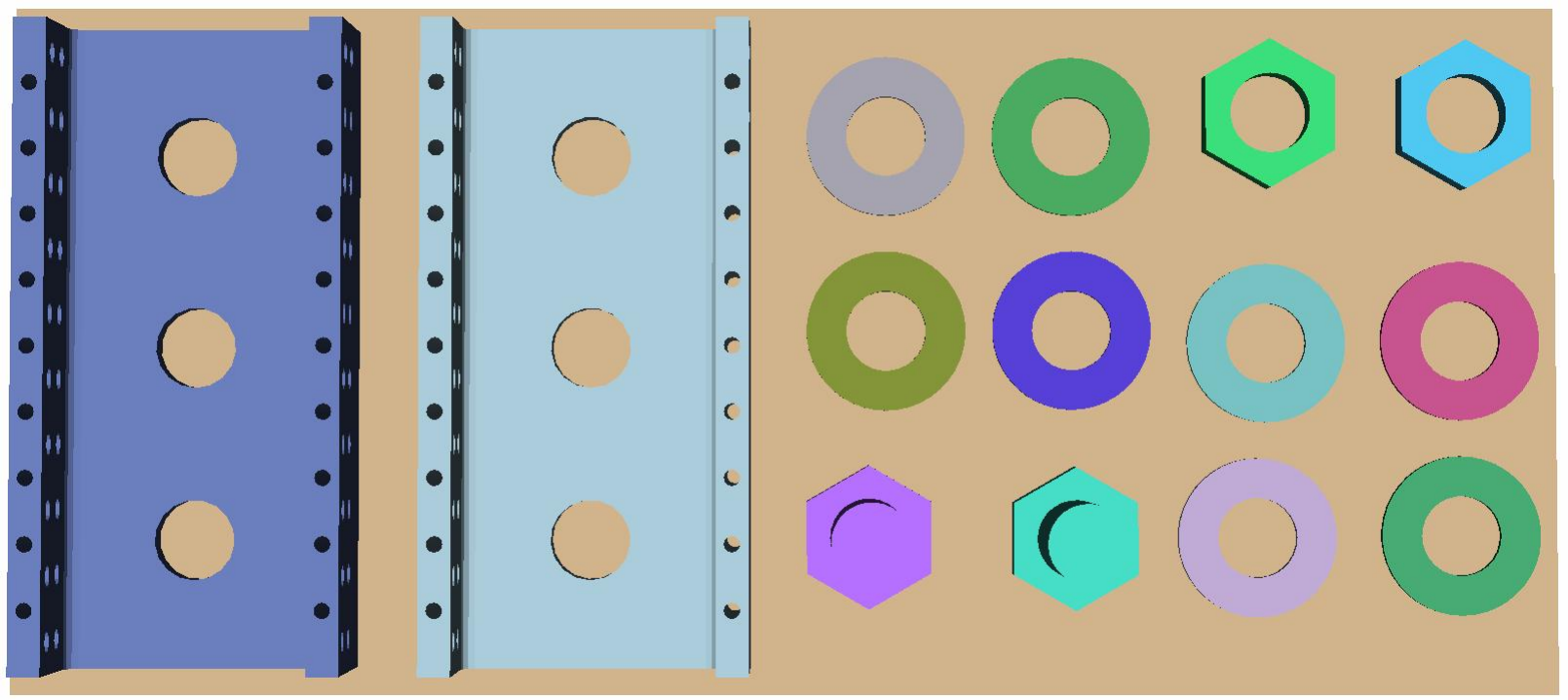}
        \caption*{Coupling Block (exploded): 14 parts}
    \end{subfigure}
    % ================== PLIERS ==================
    \begin{subfigure}[c]{\assemblyleftfrac\linewidth}
        \centering
        \includegraphics[width=\linewidth,height=3.5cm,keepaspectratio]{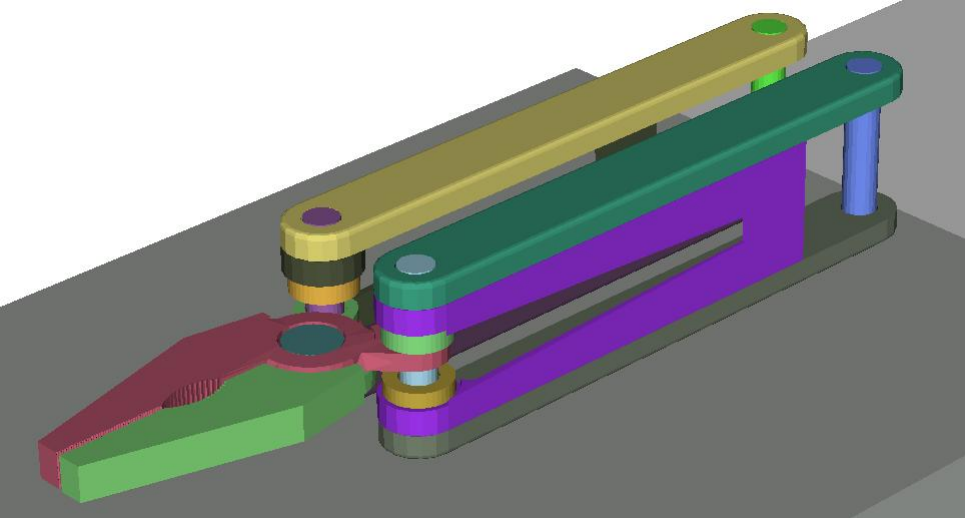}
        \caption*{Pliers (assembled)}
    \end{subfigure}
    \hfill
    \begin{subfigure}[c]{\assemblyrightfrac\linewidth}
        \centering
        \includegraphics[width=\linewidth,height=3.5cm,keepaspectratio]{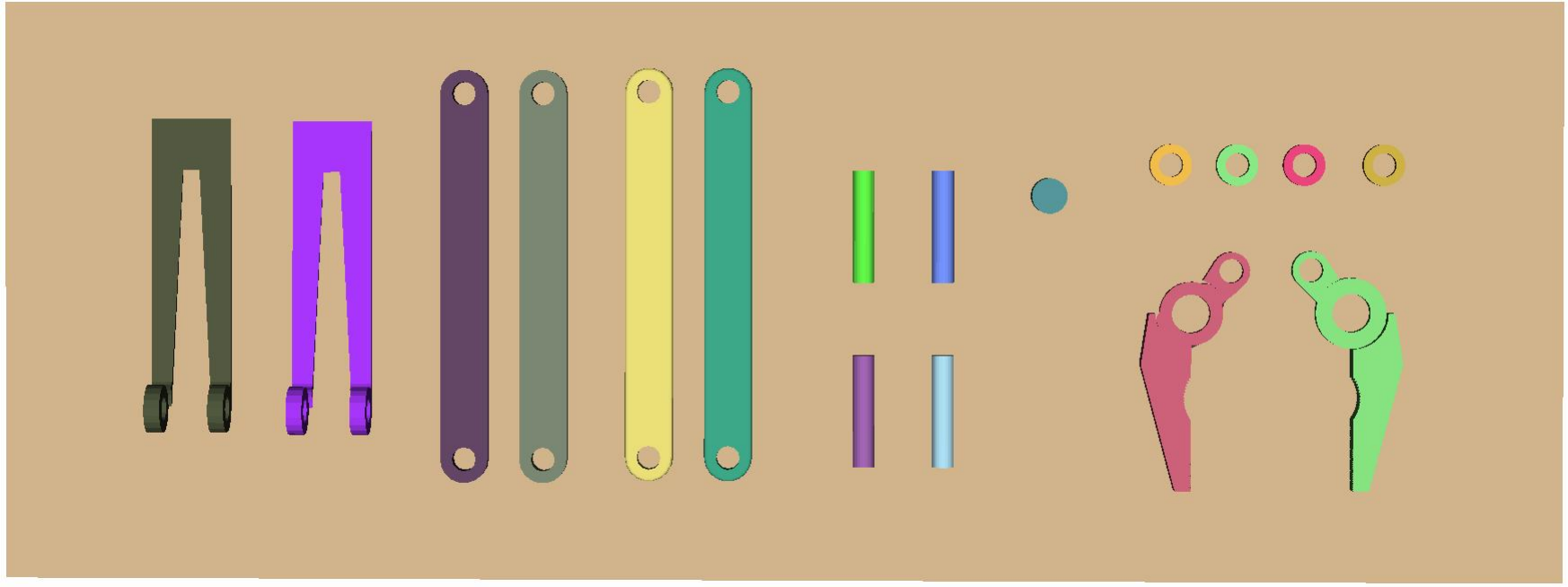}
        \caption*{Pliers (exploded): 17 parts}
    \end{subfigure}
    
    \caption{The four multi-part assemblies used for evaluation. 
      Left column: assembled state. Right column: all parts separated.}
    \label{fig:assembly_renders}
\end{figure}

To evaluate \approachNameAbbrv, we select four assemblies with 10 to 17 removable parts each:
Microscope (12 parts), Disc Brake (10 parts), Coupling Block (14
parts), and Plier (17 parts). \approachName uses a batch size of $B=10$ concurrent processes
(one CPU core per object), a per-object timeout of 180\,s, and a
batch timeout of 190\,s. For the MAB-RRT planner~\cite{bayraktar2026wafr}, we use a window size of $W=256$, an initial radius $r_0=1.0$, and a sample budget of $n=128$ for the sphere sampler. The baselines (RRT, TRRT, MateVec-TRRT)
execute objects sequentially on a single CPU core, following the
default ATA disassembly pipeline~\cite{tian2022assemble}. 

For each assembly, the scheduler randomly shuffles the initial object
queue and executes until all objects are removed or an overall timeout
of 3600\,s is reached. Each assembly is tested under 10 random initial
shuffles to evaluate robustness to ordering.
\cref{fig:assembly_renders} shows the four assemblies in their
assembled (left) and disassembled or exploded (right) states.

 \newcommand\ours{\textbf{PEEL (ours)}}
  \begin{table*}[t]
  \centering
  \caption{Environment-Specific Multi-Part Benchmark Results}
  \label{tab:multipart_results_env}
  \resizebox{\textwidth}{!}{
  \begin{tabular}{llcccccc}
  \toprule
  Environment & Planner & Success Rate (\%) & Mean Time (s) & Median Time (s) & Std Time (s) & Total Runs & Num Timeouts \\
  \midrule
  Microscope & \ours & 100.0 & 467.03 & 472.05 & 44.11 & 10 & 0 \\
   & RRT & 100.0 & 1742.87 & 1393.23 & 577.32 & 10 & 40 \\
   & TRRT & 100.0 & 914.66 & 806.65 & 511.75 & 10 & 23 \\
   & MATEVEC-TRRT & 100.0 & 818.80 & 814.62 & 378.40 & 10 & 20 \\
  \midrule
  Disc Brake & \ours & 100.0 & 188.78 & 189.49 & 27.19 & 10 & 0 \\
   & RRT & 100.0 & 1326.93 & 1100.18 & 941.97 & 10 & 36 \\
   & TRRT & 100.0 & 667.74 & 382.28 & 473.47 & 10 & 18 \\
   & MATEVEC-TRRT & 100.0 & 552.67 & 373.89 & 305.89 & 10 & 15 \\
  \midrule
  Coupling Block & \ours & 100.0 & 419.45 & 410.56 & 109.78 & 10 & 2 \\
   & RRT & 90.0 & 1865.87 & 1194.86 & 1386.14 & 10 & 64 \\
   & TRRT & 90.0 & 2120.34 & 1479.24 & 1712.07 & 10 & 72 \\
   & MATEVEC-TRRT & 100.0 & 1877.13 & 405.89 & 2446.48 & 10 & 51 \\
  \midrule
  Plier & \ours & 100.0 & 405.66 & 397.02 & 30.08 & 10 & 0 \\
   & RRT & 100.0 & 2308.05 & 1920.38 & 1356.96 & 10 & 108 \\
   & TRRT & 100.0 & 943.45 & 909.63 & 422.90 & 10 & 39 \\
   & MATEVEC-TRRT & 100.0 & 1234.75 & 736.38 & 933.60 & 10 & 40 \\
  \bottomrule
  \end{tabular}
  }
  \end{table*}

In the benchmarks, we report (1)~total disassembly time (wall-clock time from start to
complete disassembly) and (2)~overall success rate (fraction of runs
that fully disassemble the assembly within the overall timeout). 

The results are summarized in \cref{tab:multipart_results_env}. \approachNameAbbrv
achieves a 100\% success rate on all four assemblies with very few
per-object timeouts (0--2 per assembly), while the baselines accumulate
significantly more due to their sequential execution and less effective sampling in narrow passages.
In terms of mean disassembly time, \approachNameAbbrv completes the Disc
Brake assembly in 189\,s (median 189\,s), Plier in 406\,s (median
397\,s), Coupling Block in 419\,s (median 411\,s), and Microscope in
467\,s (median 472\,s). The next fastest baselines are MateVec-TRRT and TRRT, which require 2--5$\times$ longer on average, while plain RRT requires up to 7$\times$ longer.

\begin{figure*}[t]
    \centering
    \begin{subfigure}[b]{0.24\textwidth}
        \centering
        \includegraphics[width=\textwidth]{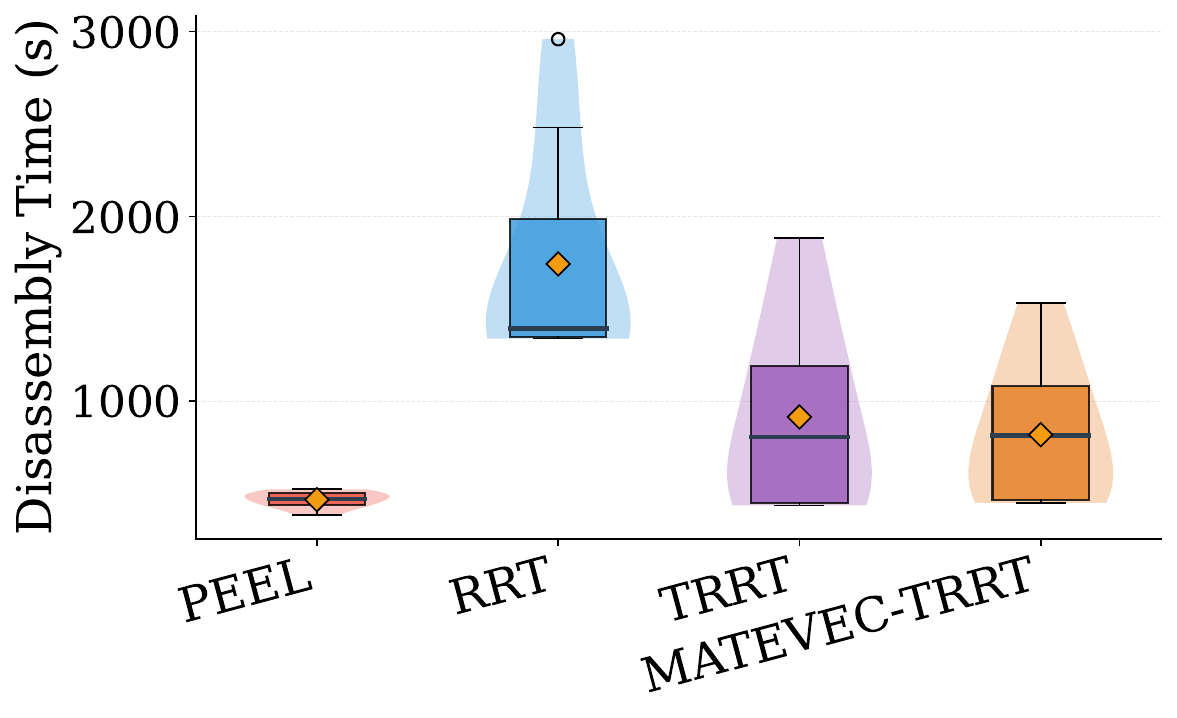}
        \caption{Microscope}
        \label{fig:runtime_00003}
    \end{subfigure}
    \begin{subfigure}[b]{0.24\textwidth}
        \centering
        \includegraphics[width=\textwidth]{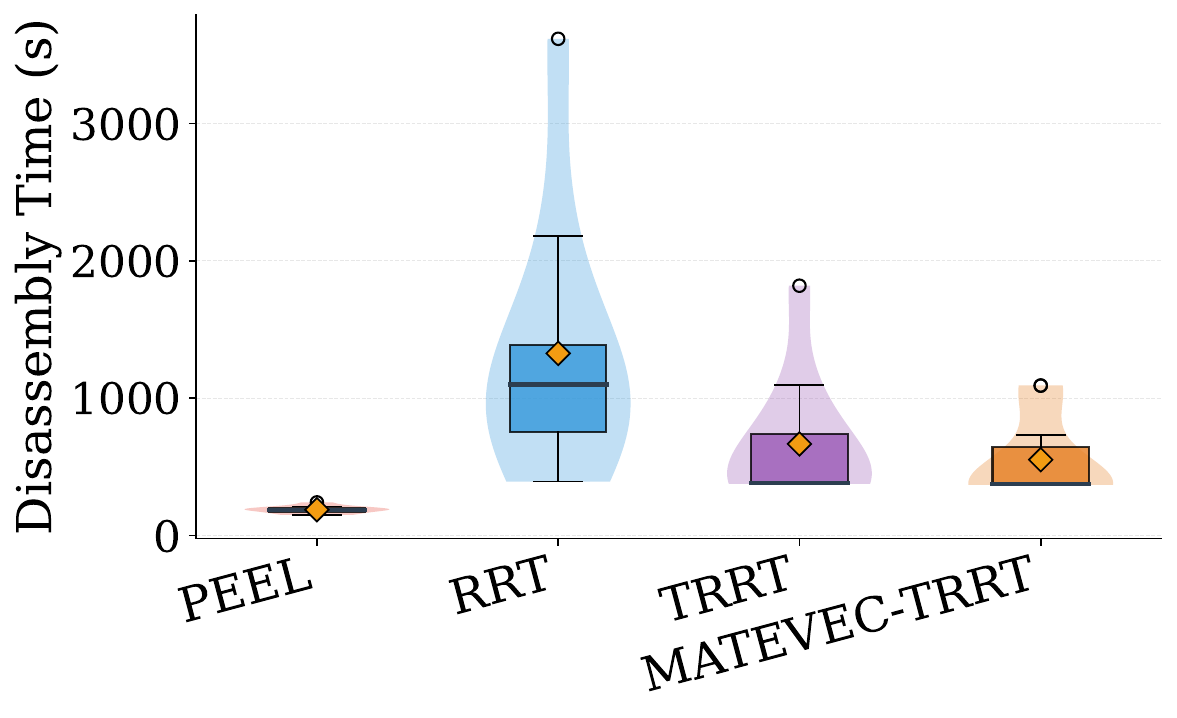}
        \caption{Disc Brake}
        \label{fig:runtime_00580}
    \end{subfigure}
    \begin{subfigure}[b]{0.24\textwidth}
        \centering
        \includegraphics[width=\textwidth]{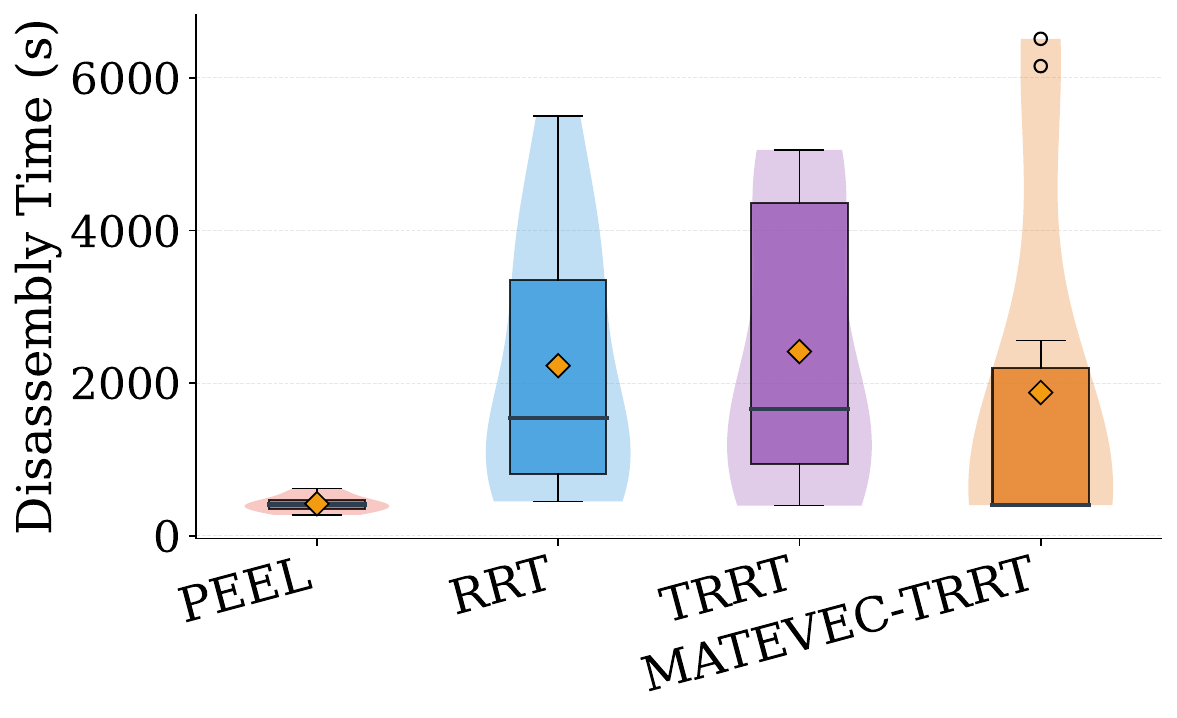}
        \caption{Coupling Block}
        \label{fig:runtime_04370}
    \end{subfigure}
    \begin{subfigure}[b]{0.24\textwidth}
        \centering
        \includegraphics[width=\textwidth]{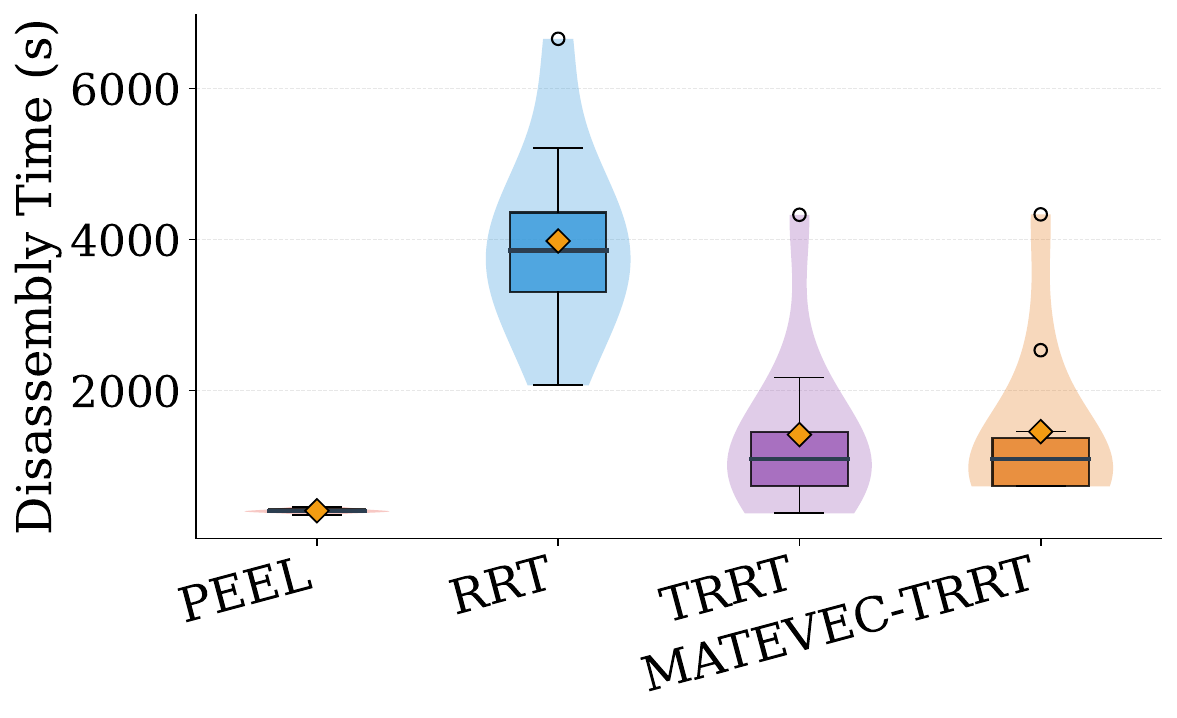}
        \caption{Pliers}
        \label{fig:runtime_zange}
    \end{subfigure}
    \caption{Total disassembly runtime comparison per assembly. Violin plots
    with overlaid box plots compare PEEL, RRT, TRRT, and MateVec-TRRT across 10
    runs each. PEEL consistently achieves an order-of-magnitude speedup with minimal variance.
    \label{fig:runtime_comparison}}
\end{figure*}

\cref{fig:runtime_comparison} shows the total disassembly time
distributions per assembly. Note that these wall-clock times reflect both
the per-object planning quality and the execution protocol: \approachNameAbbrv
benefits from parallel batching (10 cores) while the baselines run
sequentially (1 core). \approachNameAbbrv consistently clusters at the low end
of the time axis, while the baselines exhibit both higher means and
substantially larger variance. \approachNameAbbrv's standard deviation ranges from
27\,s (Disc Brake) to 110\,s (Coupling Block), compared to 306--2446\,s
for the baselines.

Coupling Block is the most challenging assembly: RRT and TRRT achieve only
90\% success rate, while MateVec-TRRT requires a mean time of 1877\,s
despite reaching 100\% success. \approachNameAbbrv solves Coupling Block in 419\,s 
mean time, demonstrating that scale-invariant sampling remains effective
even in geometrically complex multi-part settings.

\paragraph{Disassembly Progression.}

\begin{figure}[t]
      \centering
      \includegraphics[width=\columnwidth]{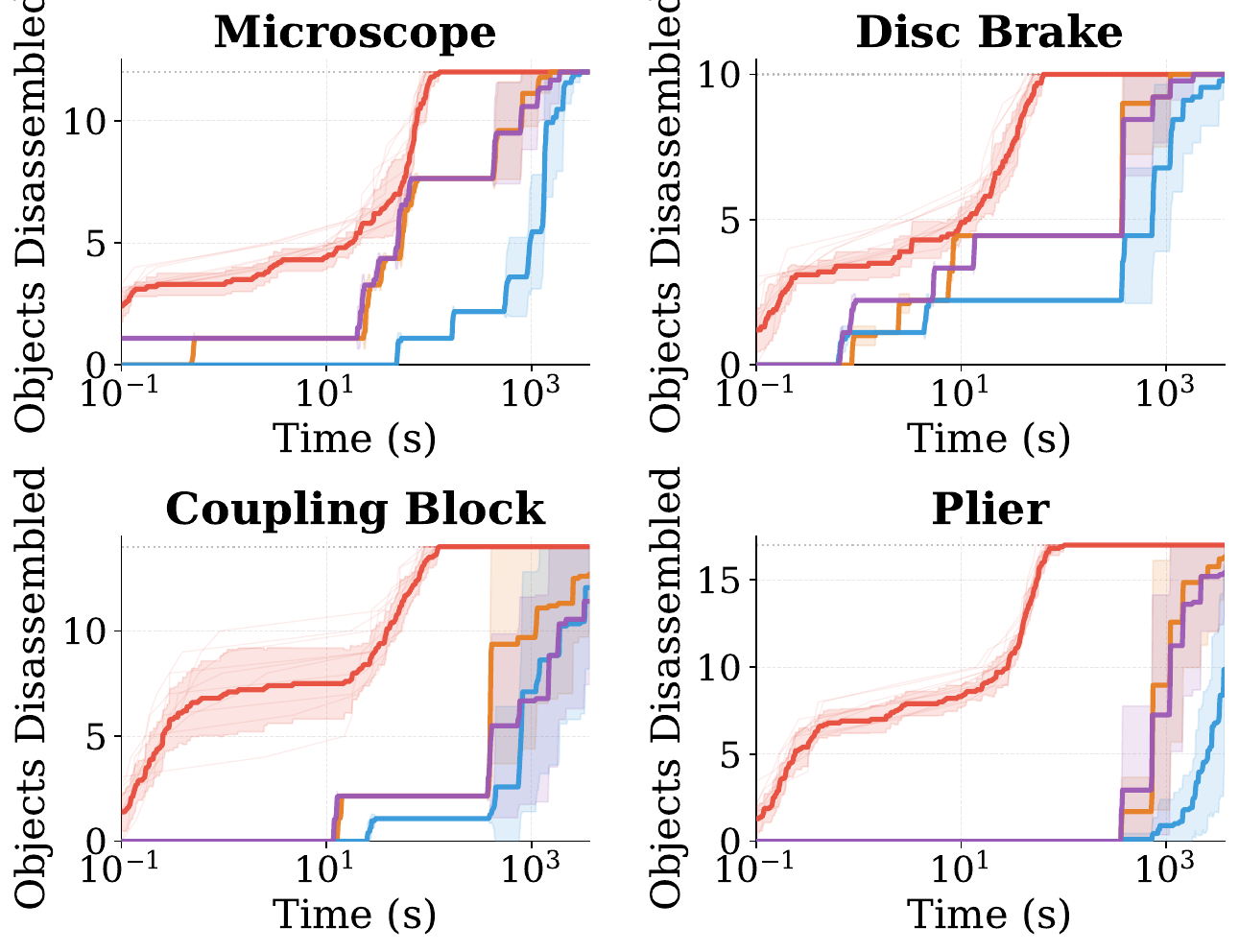}
      \caption{Objects disassembled over time (logarithmic time axis) for the
      four assemblies: \cbox{cmab}~\approachNameAbbrv, \cbox{crrt}~RRT, \cbox{ctrrt}~TRRT,
      \cbox{cmatevec}~MateVec-TRRT. \approachNameAbbrv removes objects within the first
      second and completes full disassembly one to two orders of magnitude faster
      than the baselines.}
      \label{fig:disassembly_progression}
  \end{figure}

\cref{fig:disassembly_progression} shows the number of objects
disassembled over time on a logarithmic time axis. \approachNameAbbrv (red) begins
removing objects within the first second and completes full disassembly
one to two orders of magnitude faster than the baselines. The baselines
(RRT, TRRT, MateVec-TRRT) remain near zero for the first 10--100\,s
before slowly progressing, with several runs failing to complete within
the timeout. The winner-takes-all termination strategy effectively
reduces wasted computation: once an object is successfully removed, all
concurrent planners working on other objects are immediately terminated,
allowing the system to quickly adapt to the updated geometric
constraints. The requeuing mechanism ensures that difficult objects
receive multiple attempts across successive batches as the assembly
evolves.

%%%%%%%%%%%%%%%%%%%%%%%%%%%%%%%%%%%%%%%%%%%%%%%%%%%%%%%%%%%%%%%%%%%%%%%%%
\subsection{Parameter Sensitivity\label{sec:sensitivity}}
%%%%%%%%%%%%%%%%%%%%%%%%%%%%%%%%%%%%%%%%%%%%%%%%
We assess the robustness of MAB-RRT to its main hyperparameters on the
single-object benchmark set (760 trials), varying one parameter at a time
around the default configuration ($W=256$, validity band $[10\%,50\%]$,
$r_0=1.0$, $n=128$).

\paragraph{Sliding-window size}
The UCB sliding window controls how much sampling history the bandit retains.
Across $W \in \{64,\dots,1024\}$ the planner is essentially insensitive: median
runtime stays at 4--5\,s with success at or near 100\%. Larger windows slightly
favor exploitation in tightly constrained tasks while smaller windows switch
samplers faster, but overall performance is comparable. We keep $W=256$.
\begin{table}[t]
\centering
\caption{Window size robustness statistics across 760 benchmark trials (76 models $\times$ 10 rotations). 
Fixed parameters: $n=128$, $r_0=1.0$, $r_{\max}=15.0$, validity bounds $[0.10, 0.50]$.}
\label{tab:window_size_robustness}
%\begin{tabular}{lcccccc}
\begin{tabularx}{\columnwidth}{l *{6}{>{\centering\arraybackslash}X}}
\toprule
Window Size & Mean & Median & Std Dev & Min & Max & Success Rate \\
 & (s) & (s) & (s) & (s) & (s) & (\%) \\
\midrule
$W=64$ & 5.77 & 4.37 & 4.67 & 1.14 & 32.51 & 99.7 \\
$W=256$ & 6.26 & 4.57 & 5.21 & 1.19 & 43.05 & 100.0 \\
$W=1024$ & 6.58 & 4.98 & 5.53 & 1.41 & 47.85 & 100.0 \\
\bottomrule
\end{tabularx}
\end{table}

\paragraph{Validity-rate bounds}
The target validity interval $[\alpha_{\min},\alpha_{\max}]$ sets how
aggressively the scale search shrinks or grows the sphere. Across $[5\%,15\%]$,
$[10\%,50\%]$, $[15\%,25\%]$ and $[25\%,50\%]$ all settings exceed 98\% success.
Lower bounds accept larger radii with fewer valid samples and are fastest
(median 2.63--2.68\,s), while the highest bound forces further shrinking
(4.74\,s); the resulting $1.8\times$ spread still lies in a practical range. We
use $[10\%,50\%]$.
\begin{table}[t]
\centering
\caption{MAB-RRT validity rate bounds sensitivity across 76 assemblies (760 trials).}
\label{tab:validity_sensitivity}
\begin{tabularx}{\columnwidth}{l *{4}{>{\centering\arraybackslash}X}}
\toprule
Bounds & Success Rate & Mean & Median & Runs \\
 & (\%) & (s) & (s) & \\
\midrule
{[10\%--50\%]} & 99.9 & 3.73 & 2.63 & 760 \\
{[5\%--15\%]} & 98.8 & 4.00 & 2.68 & 760 \\
{[15\%--25\%]} & 99.9 & 6.57 & 4.26 & 760 \\
{[25\%--50\%]} & 100.0 & 7.01 & 4.74 & 760 \\
\bottomrule
\end{tabularx}
\end{table}

\paragraph{Initial radius}
The initial sphere radius is corrected by the adaptive grow-shrink search, so
runtime is largely insensitive to it: $r_0 \in [0.2, 5.0]$ all converge to
3--6\,s median at near-100\% success. Only extreme values degrade, $r_0 \leq
0.1$ raises adaptation overhead and lowers success to 88--94\%, and $r_0=15.0$
keeps full success but raises median runtime to 6.2\,s. We use $r_0=1.0$.
\begin{table}[t]
\centering
\caption{Radius sensitivity statistics across 760 benchmark trials. 
Fixed parameters: $n=128$, $W=256$, validity bounds $[0.10, 0.50]$.}
\label{tab:radius_sensitivity}
\begin{tabularx}{\columnwidth}{l *{6}{>{\centering\arraybackslash}X}}
\toprule
Initial Radius & Mean & Median & Std Dev & Min & Max & Success Rate \\
 & (s) & (s) & (s) & (s) & (s) & (\%) \\
\midrule
$r_0=10^{-6}$ & 12.19 & 7.75 & 13.57 & 2.07 & 114.25 & 90.8 \\
$r_0=0.01$ & 8.72 & 4.92 & 12.73 & 1.02 & 117.77 & 88.1 \\
$r_0=0.10$ & 7.63 & 4.25 & 12.10 & 0.59 & 114.29 & 93.5 \\
$r_0=0.20$ & 5.26 & 3.40 & 6.65 & 0.56 & 98.79 & 97.8 \\
$r_0=0.50$ & 5.00 & 3.67 & 4.24 & 0.89 & 28.34 & 100.0 \\
$r_0=1.00$ & 5.61 & 4.31 & 4.48 & 1.24 & 31.85 & 100.0 \\
$r_0=2.10$ & 6.72 & 5.31 & 4.53 & 1.89 & 27.57 & 100.0 \\
$r_0=2.50$ & 7.05 & 5.23 & 6.27 & 1.66 & 51.34 & 99.9 \\
$r_0=5.00$ & 7.10 & 5.38 & 6.11 & 1.81 & 50.74 & 99.9 \\
$r_0=15.0$ & 7.81 & 6.22 & 5.80 & 2.55 & 42.16 & 100.0 \\
\bottomrule
\end{tabularx}
\end{table}

\paragraph{Sample size}
The number of quasi-random samples per sphere trades directional coverage
against validation cost. $n=128$ gives the best balance (100\% success,
2.5\,s median); $n=64$ is faster (1.7\,s) at slightly lower reliability
(99.1\%); and $n=256$ adds collision checks that slow the scale search by
roughly $1.7\times$ without improving success. We adopt $n=128$.
\begin{table}[t]
\centering
\caption{Sample size impact statistics across 760 benchmark trials. 
Fixed parameters: $r_0=1.0$, $r_{\max}=15.0$, $W=256$, validity bounds $[0.10, 0.50]$.}
\label{tab:sample_size_impact}
\begin{tabularx}{\columnwidth}{l *{6}{>{\centering\arraybackslash}X}}
\toprule
Sample Size & Mean & Median & Std Dev & Min & Max & Success Rate \\
 & (s) & (s) & (s) & (s) & (s) & (\%) \\
\midrule
$n=64$ & 2.35 & 1.69 & 2.42 & 0.36 & 39.48 & 99.1 \\
$n=128$ & 3.33 & 2.48 & 2.79 & 0.60 & 19.40 & 100.0 \\
$n=256$ & 5.58 & 4.27 & 4.39 & 1.19 & 31.01 & 100.0 \\
\bottomrule
\end{tabularx}
\end{table}

\begin{figure}[t]
    \centering
    \begin{subfigure}[b]{0.49\columnwidth}
      \centering
      \includegraphics[width=\linewidth]{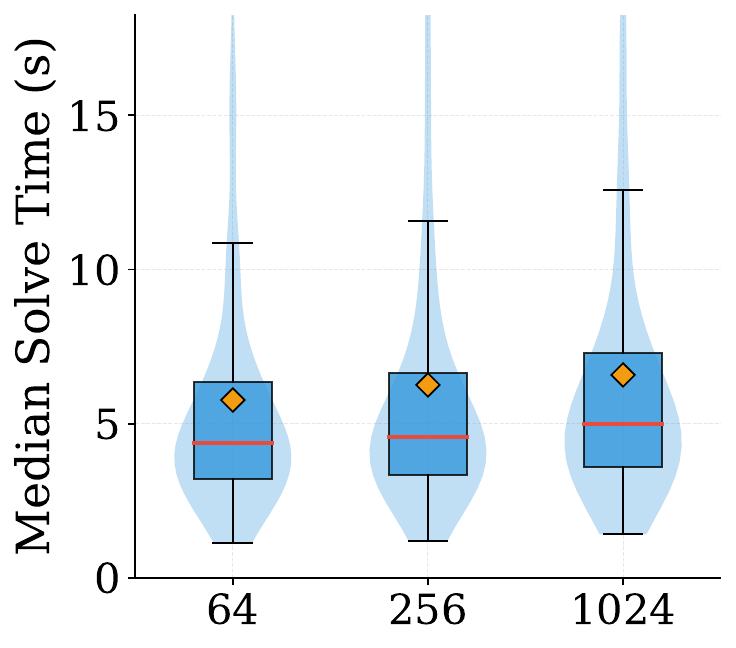}
      \caption{Sliding-window size $W$}
      \label{fig:param_window}
    \end{subfigure}
    \hfill
    \begin{subfigure}[b]{0.49\columnwidth}
      \centering
      \includegraphics[width=\linewidth]{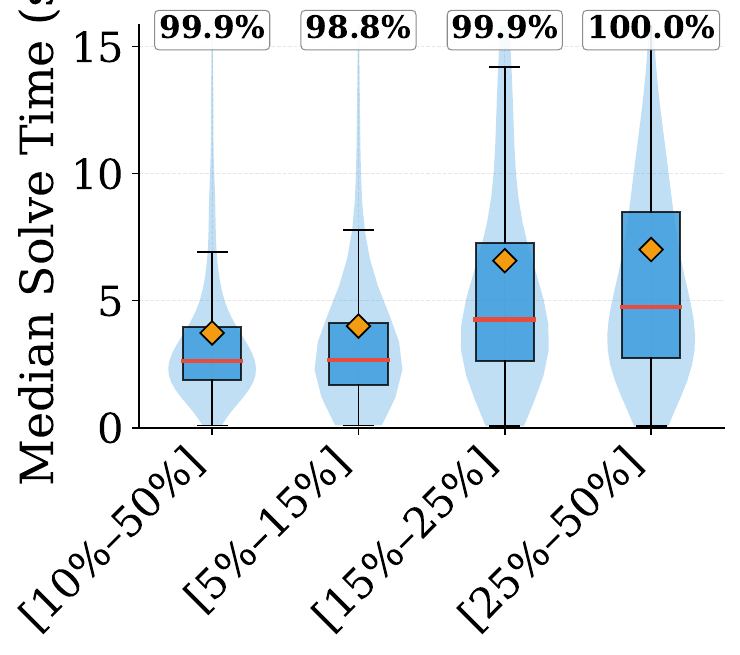}
      \caption{Validity bounds $[\alpha_{\min},\alpha_{\max}]$}
      \label{fig:param_validity}
    \end{subfigure}
    \caption{Parameter sensitivity of MAB-RRT on the single-object benchmark.}
    \label{fig:param_sensitivity}
  \end{figure}

%%%%%%%%%%%%%%%%%%%%%%%%%%%%%%%%%%%%%%%%%%%%%%%%%%%%%%%%%%%%%%%%%%%%%%%%%
\subsection{Robot Execution Demonstration\label{sec:robot_execution_demonstration}}
%%%%%%%%%%%%%%%%%%%%%%%%%%%%%%%%%%%%%%%%%%%%%%%%%%%%%%%%%%%%%%%%%%%%%%%%%
\begin{figure*}[!t]
    \centering
    \begin{subfigure}{\textwidth}
        \centering
        \includegraphics[width=\textwidth]{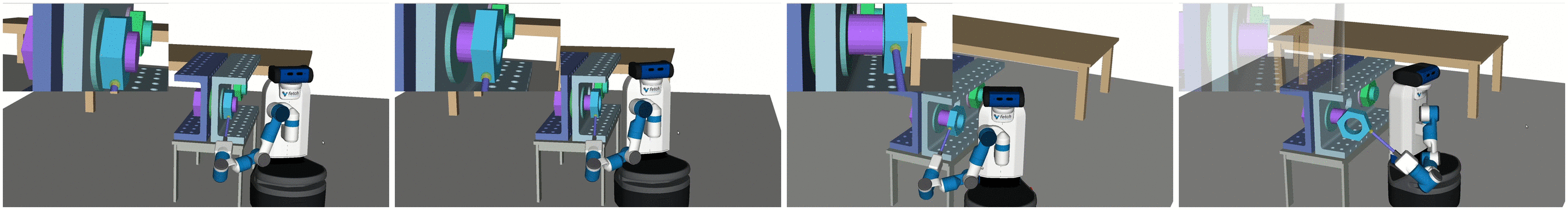}
        \caption{Coupling Block}
        \label{fig:robot_execution_coupling_block}
    \end{subfigure}
    \par\smallskip
    \begin{subfigure}{\textwidth}
        \centering
        \includegraphics[width=\textwidth]{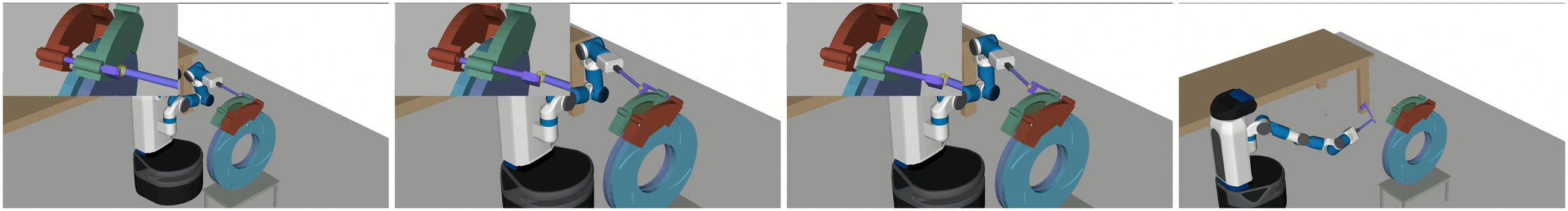}
        \caption{Disc Brake}
        \label{fig:robot_execution_disc_brake}
    \end{subfigure}
    \par\smallskip
    \begin{subfigure}{\textwidth}
        \centering
        \includegraphics[width=\textwidth]{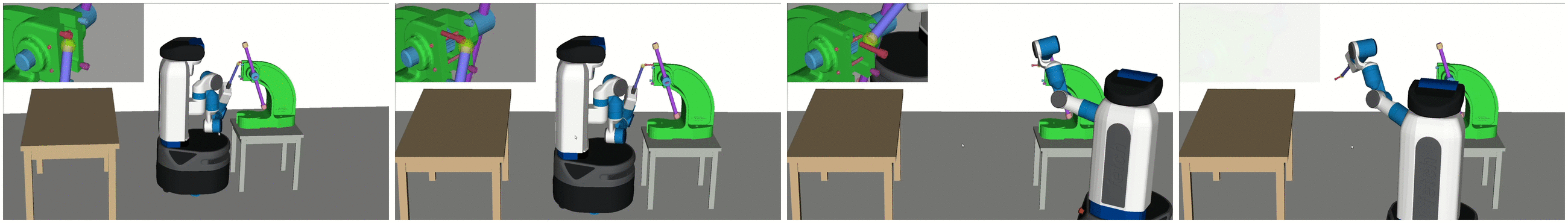}
        \caption{Microscope}
        \label{fig:robot_execution_microscope}
    \end{subfigure}
    \par\smallskip
    \begin{subfigure}{\textwidth}
        \centering
        \includegraphics[width=\textwidth]{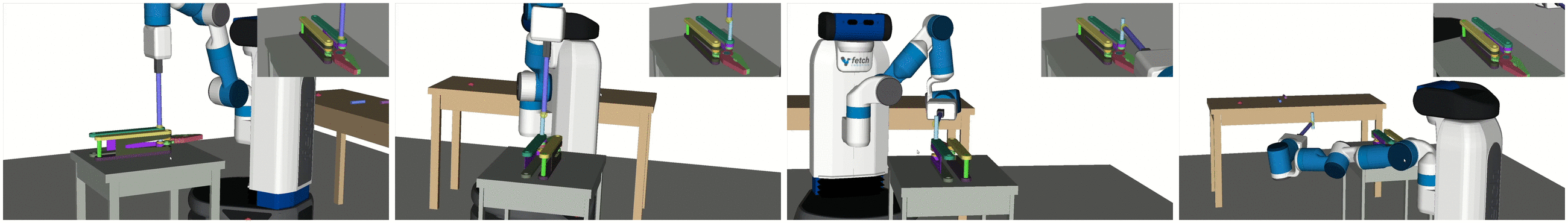}
        \caption{Plier}
        \label{fig:robot_execution_plier}
    \end{subfigure}
    \caption{Robot demonstrations using the robot execution pipeline. The disassembly sequences are computed using PEEL and individual escape paths computation uses scale-invariant sampling inside MAB-RRT. All sequences were successfully executed using the Fetch robot.}
    \label{fig:robot_demonstrations}
\end{figure*}

% \begin{figure*}[!t]
%     \centering
%     \begin{subfigure}{\textwidth}
%         \centering
%         \includegraphics[width=\textwidth]{images/robot_demonstrations/coupling_block_grid.png}
%         \caption{Coupling Block}
%         \label{fig:robot_execution_coupling_block}
%     \end{subfigure}
%     \hfill
%     \begin{subfigure}{\textwidth}
%         \centering
%         \includegraphics[width=\textwidth]{images/robot_demonstrations/disc_brake_grid.png}
%         \caption{Disc Brake}
%         \label{fig:robot_execution_disc_brake}
%     \end{subfigure}
%     \hfill
%     \begin{subfigure}{\textwidth}
%         \centering
%         \includegraphics[width=\textwidth]{images/robot_demonstrations/microscope_grid.png}
%         \caption{Microscope}
%         \label{fig:robot_execution_microscope}
%     \end{subfigure}
%     \hfill
%     \begin{subfigure}{\textwidth}
%         \centering
%         \includegraphics[width=\textwidth]{images/robot_demonstrations/plier_grid.png}
%         \caption{Pair of Pliers}
%         \label{fig:robot_execution_plier}
%     \end{subfigure}
%     \caption{Robot demonstrations using the robot execution pipeline. The disassembly sequences are computed using PBP and individual escape paths computation uses scale-invariant sampling inside MAB-RRT. All sequences were successfully executed using the Fetch robot.}
%     \label{fig:robot_demonstrations}
% \end{figure*}

After showing that \approachNameAbbrv can successfully generate object sequences to be disassembled, we demonstrate next how those sequences can be
translated into feasible robot motions. To validate this, we deployed
the five-phase execution protocol
(\cref{sec:robot_pipeline}) on all four multi-part assemblies in
DARTSim~\cite{lee2018dart} via Robowflex~\cite{kingston2022robowflex}.

The robot is a simulated Fetch mobile manipulator with 11 degrees of
freedom: a 7-DOF arm, a 1-DOF torso lift, and a 3-DOF mobile base. The
end-effector is equipped with a spherical grasping tool that simulates
stable point contacts on arbitrarily shaped surfaces. During extraction
and transport, the grasped object is rigidly attached to the
end-effector via a weld joint, which eliminates the need for grasp
stability analysis and ensures the object cannot slip during motion.
Collision checking remains active throughout all phases: every candidate
robot configuration is validated against the full environment, including
the assembly, other objects, and the table.

\paragraph{Parameters}

The following parameters are specific to the five-phase online execution
protocol (\cref{sec:robot_pipeline}); offline \approachNameAbbrv and MAB-RRT settings are
as reported in \cref{sec:multi_part_benchmarks,sec:sensitivity}.

\begin{itemize}
    \item \textbf{PCA cylinder extension factor} ($\delta$): Scales the
    principal-component sampling cylinder as
    $h_{\text{ext}} = \delta \cdot r^\star$ during offline MAB-RRT path
    generation.

    \item \textbf{Grasp surface sample budget}: Maximum number of random
    surface points drawn from each object mesh when searching for a valid
    grasp in Phase~I (and during intermediate regrasps in Phase~II).
    Value: $5000$.

    \item \textbf{Grasp sampling margin} ($\mu_g$): Offset from the
    object surface at which candidate end-effector poses are placed
    before IK and collision validation (Phases~I and~III).

    \item \textbf{IK seed configuration}: Previous accepted robot
    configuration used to initialize inverse kinematics at each object
    waypoint in Phase~II, encouraging smooth joint trajectories.

    \item \textbf{Mobility perturbation magnitude} ($\epsilon_t$):
    Displacement size used to probe translational and rotational object
    mobility when detecting the free-state boundary in Phase~II.

    \item \textbf{Goal grasp sample budget} ($N_g'$): Maximum number of
    candidate goal grasps sampled and validated in Phase~III before
    reporting failure.

\end{itemize}

\paragraph{Results}

All four assemblies (Microscope, Coupling Block, Disc Brake, and Plier)
were fully disassembled in simulation. The regrasp mechanism proved
essential for parts requiring long extraction motions that exceeded the
manipulator's workspace from a single grasp configuration. The
free-state detection successfully identified the transition from
constrained to unconstrained motion in each extraction, allowing the
transport phase to plan freely rather than follow the remainder of the
narrow-passage trajectory. The demonstrations confirm that the geometric extraction paths computed by MAB-RRT can be executed on a kinematic robot: the constrained, narrow-passage portion is followed directly through inverse kinematics, while regrasping absorbs kinematic and workspace limits and the remaining free-space transport is replanned. Individual frames of the robot disassembly             
sequences are shown in \cref{fig:robot_demonstrations} including 
the microscope (\cref{fig:robot_execution_microscope}), 
the disc brake (\cref{fig:robot_execution_disc_brake}), 
the coupling block (\cref{fig:robot_execution_coupling_block}), 
and the pair of pliers (\cref{fig:robot_execution_plier}). 
Full videos are available on our website\footnote{\url{https://peel-disassembly.surge.sh/}}.
\section{Conclusion\label{sec:discussion-and-limitations}}

We presented the \approachName (\approachNameAbbrv), a framework that computes geometrically feasible multi-part disassembly sequences and executes them on a robot manipulator. \approachNameAbbrv discovers removal orders through concurrent single-object planning races, each solved by MAB-RRT with scale-invariant sampling to separate parts through tight, narrow passages, and avoids the exponential precedence search of combinatorial sequencing. We demonstrated a 100\% success rate on 76 single-object benchmarks and on four multi-part assemblies (Microscope, Disc Brake, Coupling Block, and Plier, 10 to 17 parts
each), achieving the lowest runtime against three baselines, and we showed that the resulting sequences are executable by a simulated Fetch manipulator through the five-phase execution pipeline.

While the results are promising, two important opportunities remain for strengthening this approach and pushing our work towards real world deployment. Particularly, those are:

\begin{itemize}
  \item \textbf{Physical Simulation}: Our framework focuses on geometrically feasible paths to achieve a disassembly state. We believe this is an important and necessary step towards the efficient disassembly of real-world objects. Future work will concentrate on using geometrically feasible paths from \approachNameAbbrv as heuristics to guide manipulation systems towards dynamically feasible paths.  
  \item \textbf{Non-sequential Disassembly}: Our system currently handles objects which are sequentially decomposable. However, some objects are kinematically interlocked~\cite{zhang2021interlocking,tian2022assemble} (like the Burr puzzle~\cite{xin2011burrpuzzles}) meaning that two or more objects have to be moved \emph{simultaneously} to achieve a disassembly state. A straightforward extension of \approachNameAbbrv would involve running MAB-RRT on combinations of pairs (or higher order tuples) if no single-object solutions are found after some time.
\end{itemize}

Despite those open opportunities, \approachNameAbbrv already provides a general-purpose framework for disassembly problems and we have shown that the resulting paths are executable with a manipulator robot. We therefore believe we have made a significant step towards general-purpose disassembly frameworks for robots in the real-world.

\appendices              
 \section{Planner Configurations}
  \label{app:configs}

  This appendix documents the exact planner configuration behind every result.
  \Cref{tab:app_default} gives the recommended MAB-RRT configuration, grouped by
  role. Every listed key is read and used by the planner (\texttt{loadYAMLConfig}
  in \texttt{MAB\_RRT.cpp}); the released configuration files are byte-faithful to
  the runs and record the source file and its checksum. Most studies share this
  base, but a few were executed on a different one; \Cref{tab:app_perexp} therefore
  lists the configuration actually used for each results table, so every number is
  traceable to its exact settings.

  \begin{table}[t]
  \centering
  \footnotesize
  \setlength{\tabcolsep}{4pt}
  \caption{Recommended MAB-RRT configuration.}
  \label{tab:app_default}
  \begin{tabular}{@{}llr@{}}
  \toprule
  Group & Parameter & Value \\
  \midrule
  \multirow{2}{*}{Goal bias}
    & uniform arm goal bias & $0.5$ \\
    & PCA arm goal bias & $10^{-7}$ \\
  \addlinespace
  \multirow{2}{*}{Bandit}
    & sliding-window size $W$ & $256$ \\
    & forced-uniform streak & $5$ \\
  \addlinespace
  \multirow{8}{*}{Scale search}
    & sample budget $n$ & $128$ \\
    & initial radius $r_0$ & $1.0$ \\
    & radius clamp $[r_{\min},r_{\max}]$ & $[10^{-6},\,15.0]$ \\
    & shrink factor $s$ & $\exp(-0.7)$ \\
    & grow factor $g$ & $\exp(0.9)$ \\
    & validity band $[\alpha_{\min},\alpha_{\max}]$ & $[0.10,\,0.50]$ \\
    & max burn-in steps & $50$ \\
    & Fibonacci-lattice jitter & $\pi/8$ \\
  \addlinespace
  \multirow{4}{*}{PCA sampler}
    & online PCA recompute & on \\
    & cylinder radius offset mult. & $0.1$ \\
    & cylinder sampling radius mult. & $1.5$ \\
    & height extension $\varepsilon$ & $2.0$ \\
  \addlinespace
  \multirow{3}{*}{Pre-check / exit}
    & free-sampling probability & $0.20$ \\
    & uniform pre-check trials & $0$ \\
    & burn-in / full-validity early exit & off \\
  \addlinespace
  \multirow{2}{*}{Rewards}
    & uniform valid / invalid & $10^{7}$ / $0$ \\
    & PCA valid / invalid & $1.0$ / $0$ \\
  \bottomrule
  \end{tabular}
  \end{table}

  \begin{table}[t]
  \centering
  \footnotesize
  \setlength{\tabcolsep}{4pt}
  \caption{Per-experiment configuration behind each results table. The parameter
  varied by a sensitivity study is marked \emph{var.}; all others are held at the
  listed value. Shrink / grow factors are $\exp(-0.7)$ / $\exp(0.9)$ throughout.}
  \label{tab:app_perexp}
  \begin{tabular}{@{}lcccc@{}}
  \toprule
  Experiment & $W$ & $n$ & $r_0$ & $[\alpha_{\min},\alpha_{\max}]$ \\
  \midrule
  Single-object benchmark & $256$ & $128$ & $1.0$ & $[0.10,0.50]$ \\
  Window size            & \emph{var.} & $256$ & $1.0$ & $[0.10,0.50]$ \\
  Initial radius         & $256$ & $256$ & \emph{var.} & $[0.10,0.50]$ \\
  Sample size            & $256$ & \emph{var.} & $1.0$ & $[0.10,0.50]$ \\
  Validity band          & $128$ & $128$ & $1.0$ & \emph{var.} \\
  Multi-part             & $256$ & $128$ & $1.0$ & $[0.10,0.50]$ \\
  \bottomrule
  \end{tabular}
  \end{table}

  The per-trial timeout is $120$\,s for all single-object studies. The multi-part
  study additionally uses a batch size of $B=10$ concurrent planners, a per-object
  timeout of $180$\,s, a batch timeout of $190$\,s, and an overall run timeout of
  $3600$\,s.

  \subsection{Sensitivity sweeps}
  \label{app:sweeps}

  \Cref{tab:app_sweeps} lists the value set explored by each sensitivity study.
  Each study varies the single axis shown while the remaining parameters stay at
  the per-experiment values in \Cref{tab:app_perexp}.

  \begin{table}[t]
  \centering
  \footnotesize
  \setlength{\tabcolsep}{4pt}
  \caption{Parameter-sensitivity sweeps. One axis varies per study.}
  \label{tab:app_sweeps}
  \begin{tabular}{@{}lll@{}}
  \toprule
  Study & Axis & Values \\
  \midrule
  Window size & $W$ & $\{64,\,256,\,1024\}$ \\
  Initial radius & $r_0$ & $10^{-6},0.01,0.1,0.2,0.5,$ \\
   & & $1.0,2.1,2.5,5.0,15.0$ \\
  Sample size & $n$ & $\{64,\,128,\,256\}$ \\
  Validity band & $[\alpha_{\min},\alpha_{\max}]$ & $[0.05,0.15],[0.10,0.50],$ \\
   & & $[0.15,0.25],[0.25,0.50]$ \\
  \bottomrule
  \end{tabular}
  \end{table}

  \subsection{Multi-part free-state condition}
  \label{app:freestate}

  The multi-part pipeline reuses the default planner configuration and adds a
  per-assembly free-state test (Phase~II termination): a part is free once its
  translational and rotational mobility ranks reach $(\rho_t,\rho_r)$, probed with
  step sizes $(\epsilon_t,\epsilon_r)$. \Cref{tab:app_freestate} gives the values;
  assemblies not listed use the default $(\rho_t,\rho_r)=(2,2)$.

  \begin{table}[t]
  \centering
  \footnotesize
  \setlength{\tabcolsep}{4pt}
  \caption{Per-assembly free-state thresholds (multi-part pipeline).}
  \label{tab:app_freestate}
  \begin{tabular}{@{}llll@{}}
  \toprule
  Assembly & $(\rho_t,\rho_r)$ & $\epsilon_t$ & $\epsilon_r$ \\
  \midrule
  Microscope (00003) & $(2,2)$ & $0.15$\,m & $10^{\circ}$ \\
  Coupling Block (04370) & $(3,2)$ & $0.08$\,m & $6^{\circ}$ \\
  Disc Brake (00580), Plier (zange) & $(2,2)$ & $0.10$\,m & $7^{\circ}$ \\
  \bottomrule
  \end{tabular}
  \end{table}

  \subsection{Robot execution pipeline}
  \label{app:robot}

  The five-phase execution protocol adds the parameters in
  \Cref{tab:app_robot}; the offline \approachNameAbbrv and MAB-RRT settings are unchanged from
  \Cref{tab:app_default,tab:app_perexp}. The free-state thresholds
  $(\rho_t,\rho_r)$ and probe sizes $(\epsilon_t,\epsilon_r)$ are per-assembly and
  are given in \Cref{tab:app_freestate}; the cylinder height extension is the
  MAB-RRT default ($\varepsilon=2.0$).

  \begin{table}[t]
  \centering
  \footnotesize
  \setlength{\tabcolsep}{4pt}
  \caption{Robot execution pipeline parameters (five-phase protocol).}
  \label{tab:app_robot}
  \begin{tabular}{@{}llr@{}}
  \toprule
  Phase & Parameter & Value \\
  \midrule
  \multirow{3}{*}{I: Grasp}
    & grasp sampling margin $\mu_g$ & $0.05$\,m \\
    & grasp surface sample budget & $5000$ \\
    & grasp resample rounds & $10$ \\
  \addlinespace
  \multirow{2}{*}{III: Goal IK}
    & goal-grasp resample rounds $N_g'$ & $5$ \\
    & goal / bridge IK attempts per round & $100$ / $100$ \\
  \addlinespace
  \multirow{3}{*}{Planners}
    & approach (Phase~I) & RRTConnect \\
    & bridge (Phase~IV) & RRTConnect \\
    & transport (Phase~V) & RRTConnect \\
  \bottomrule
  \end{tabular}
  \end{table}

  The full configuration files are released with the code (see
  \texttt{reproducibility/README.md}).

\bibliographystyle{IEEEtranS}
{
\balance
%\small
\bibliography{general}
}
\end{document}